%% file: thesis.tex
\providecommand{\main}{.}
\documentclass[12pt]{report}          

\usepackage{bm}
\usepackage{subfig}

\usepackage[utf8]{inputenc}
\usepackage{subfiles}
\usepackage[titletoc]{appendix}
\usepackage{amssymb,amsmath}
\usepackage{url}
\usepackage{nameref,zref-xr}
\zxrsetup{toltxlabel}
\usepackage[breaklinks=true,hidelinks]{hyperref} 
\usepackage{tabulary} 
\usepackage{rotating} 
\usepackage{multirow} 
\usepackage{graphicx}
\usepackage{enumitem} 
\usepackage{mdwlist}	
\usepackage{xspace}
\usepackage{xcolor}
\usepackage{csquotes} 
\usepackage{xmpincl} 

\usepackage[%
backend=biber,%
style=ieee,%
backref=true,%
sortcites=true,sorting=nyt,%
mincitenames=1,maxcitenames=2%
]{biblatex}

\usepackage[american]{babel}

\renewbibmacro*{pageref}{%
   \iflistundef{pageref}
     {}
     {%
      \printtext{\addperiod\hfill\rlap{\hskip15pt\colorbox{blue!5}{\scriptsize\printlist[pageref][-\value{listtotal}]{pageref}}}}}}

\usepackage[nogroupskip,nonumberlist,nopostdot]{glossaries}

\setacronymstyle{long-short-desc}
\setglossarystyle{altlist}

\makenoidxglossaries 

\loadglsentries{\main/tex/glossary}

\usepackage{\main/uathesis}  

\makeatletter
\DeclareRobustCommand\onedot{\futurelet\@let@token\@onedot}
\def\@onedot{\ifx\@let@token.\else.\null\fi\xspace}

\makeatother

\renewcommand{\contentsname}{Table of Contents}

\title{An Empirical Study on Learning and Improving the Search Objective for Unsupervised Paraphrasing} 
\author{Weikai Steven Lu}

\degree{\MSc}

\dept{Computing Science}  

\submissionyear{2022}

\newcommand{\onlyinsubfile}[1]{#1}
\newcommand{\notinsubfile}[1]{}

\begin{document}

\renewcommand{\onlyinsubfile}[1]{}
\renewcommand{\notinsubfile}[1]{#1}

\preamblepagenumbering 
\titlepage 

\subfile{\main/tex/abstract.tex}



%

\subfile{\main/tex/dedication.tex}

\subfile{\main/tex/quote.tex}

\truedoublespacing

\subfile{\main/tex/acknowledgements.tex}

\singlespacing 
               
\tableofcontents
\listoftables  
\listoffigures 

               
               
\truedoublespacing

\setforbodyoftext 


\subfile{\main/tex/introduction.tex}

\subfile{\main/tex/background.tex}

\subfile{\main/tex/methodology.tex}

\subfile{\main/tex/experiments.tex}

\subfile{\main/tex/conclusion.tex}

\renewcommand\bibname{References}
\clearpage\addcontentsline{toc}{chapter}{\bibname}
%
\singlespacing 
%
\printbibliography

\doublespacing

%


\end{document}

%% file: tex/glossary.tex
\longnewglossaryentry{sampleGlossaryEntry}{
    name={Glossary Entry},
    text={glossary entry}
}{
}

\newacronym[description={A sample acroynm description}]{asa}{ASA}{A Sample Acronym}

%% file: tex/abstract.tex
\begin{abstract}
Research in unsupervised text generation has been gaining attention over the years. One recent approach is local search towards a heuristically defined objective, which specifies language fluency, semantic meanings, and other task specific attributes. Search in the sentence space is realized by word-level edit operations including insertion, replacement, and deletion. However, such objective function is manually designed with multiple components. Although previous work has shown maximizing this objective yields good performance in terms of true measure of success (i.e. BLEU and iBLEU), the objective landscape is considered to be non-smooth with significant noises, posing challenge for optimization.

In this dissertation, we address the research problem of smoothing the noise in the heuristic search objective by learning to model the search dynamics. Then, the learned model is combined with the original objective function to guide the search in a bootstrapping fashion.

Experimental results show that the learned models combined with the original search objective can indeed provide a smoothing effect, improving the search performance by a small margin.
\end{abstract}

%% file: tex/dedication.tex
\begin{dedication}
	\vspace*{1in}
	\begin{center}
	         \emph{To my parents and my brother} \\
             \emph{For always giving me unconditional love and support.}
	\end{center}
\end{dedication}

%% file: tex/quote.tex
\begin{quotepage}
 \vspace*{1in}
 \begin{center}
	\emph{The computer was born to solve problems that didn't exist before.}
	\begin{flushright}
		-- Bill Gates.
	\end{flushright}
 \end{center}
\end{quotepage}

%% file: tex/acknowledgements.tex
\begin{acknowledgements} 
First and formost, I would like to thank my supervisor Professor Lili Mou for his valuable guidance and feedback throughout my research. His insight and knowledge into the subject matter was invaluable in helping me formulating the research questions and methodology, as well as pushing me to sharpen my thinking and brought my work to a higher level. I would also like to pay my regards to my committee Dr. Alona Fyshe and Dr. Davood Rafiei for providing their valuable suggestion for my thesis.

I would like to thank everyone who supported me during my master program at the University of Alberta. I would like to thank Professor Martin Jagersand, whom I was fortunate to work with during my undergraduate study at the University of Alberta before entering the master program. He introduced me into research in computing science, and provided me with opportunities and horizons that I needed to choose the right direction in my research. I would like to thank all the friends I made during my two years of master program at University of Alberta who have shared with me priceless experiences and stories: Anup Anand Deshmukh, Chen Jiang, Chenyang Huang, Jiabin Fan. Special thanks to Mu He for being my best friend in life and always providing me with the warmest support.

I am deeply grateful to Wyatt Praharenka. Wyatt and I spent numerous late nights on campus combating school works. The support we have for each other helped us through innumerable down times, and we are now both graduating with our master's degrees.

Finally, I would like to express my utmost gratitude and appreciation for my family. Thanks to my mother Muyi, for all her selfless love and caring throughout the years. Thanks to my father, Jianjun, for his invaluable life experience and support. Thanks to my brother, Tony, for all the good memories we had together. 
\end{acknowledgements}

%% file: tex/introduction.tex
\chapter{Introduction}
\section{Background}
    Natural language generation (NLG) has been a long standing task in the field of Natural Language Processing (NLP) over the years. Being able to understand, model and generate texts that are syntactically compliant, semantically meaningful, coherent with contexts, and free from rigid rules and artificial construction remains a challenge in NLP research. 

    A wide range of applications are dependent on text generation modules including dialogue systems, data-to-text generation, document summarization, sentence simplification, style transfer, and machine translation. 
    
    In this dissertation, we focus on conditional text generation, whose goal is to generate a piece of text based on a given piece of source text, while enforcing some task specific attributes. For example, the desired attributes for paraphrase generation task would be preserving semantic meaning from the input text, while using different wording; for machine translation, the generated text is required to deliver the same semantic meaning as the input text, while being syntactically coherent.
    
	Due to the complexity of language generation tasks, traditional approaches rely on rules and templates. However, modern languages evolve over time through the pragmatic use by humans without conscious planning and premeditation \cite{lyons1991natural}. In other words, natural languages used by humans are fluid in terms of grammar and structures. For this reason, rules and templates are oftentimes restricted to narrow applications: for example, rules and templates used by a hotel booking chatbot are very different from that of an internet technical support service agent, leading to the need of sophisticated task-specific design for each application. Despite the high efficiency and controllability, rules and templates are never able to generate true natural language that are diverse, complex, while being able to deliver important information.
	
	Recent advances in computational hardware and large scale machine learning models have revolutionized the paradigm of text generation. State-of-the-art language models powered by Artificial Neural Networks (ANN) have demonstrated their superior capability of modeling complex environmental states in various tasks. Language models powered by ANNs can now learn and represent complex natural languages as a black-box by learning from a large amount of training data.
	Such a data-driven approach to model languages eliminates the need of static rules and templates hand-crafted by humans, as the languages are modeled implicitly by the neural networks learning from true natural language data directly. Being able to model natural languages in such a free form has enabled numerous breakthroughs in NLP.
	
	The most widely used approach for supervised text generation nowadays is the neural network-based sequence-to-sequence (Seq2Seq) model \cite{NIPS2014_a14ac55a}. The role of Seq2seq model is to produce a mapping between sequential data. A Seq2Seq model first encodes the language input into a numerical representation using a neural network-based encoder. Such representation carries the semantic meaning and structural information of the input sentence, which is then used to generate output text by the decoder. With parallel corpora available, training of such a model is typically done by maximizing the probability of generating the ground-truth output given the input. Various modules are proposed to parameterize this encoding-decoding process, including recurrent neural networks (RNN), and the more advanced Transformer \cite{transformer} architecture with attention mechanisms.
	
	However, parallel corpora are labor-intensive to acquire, and many applications do not come with a massive amount of parallel training data. For example, many rare languages or tasks are not well studied yet in the field of NLP, thus simply do not have large amount of annotated data for training supervised models. Another common scenario is when a model is to be used in a new domain, all training corpora dedicated for the old application is no longer usable, which is often the case when a company needs to quickly develop a minimum viable product for new applications. Being able to generate text without supervision by parallel corpora is very much in need in a lot of scenarios.
    
    Unsupervised text generation has attracted increasing interest in the field of NLP over the years. One intuitive approach is the variational auto-encoder \cite{bowman-etal-2016-generating}, which generates text by sampling from a latent space. However, low interpretability and controllability of the latent space pose challenges on controlling the text being generated. Consider the example of sentence simplification, where the goal is to shorten a sentence while preserving the original content. This goal would not be trivial to specify in the latent space, as it would not be apparent how to modify the latent representation would shorten the sentence length without compromising the content.
    
	Another approach to unsupervised text generation is stochastic search towards a heuristically defined objective function. The objective function measures the quality of generated text in terms of general attributes including language fluency, expression fluency, and other task specific attributes such as length and expression diversity. Then the goal of the search algorithm is to find an output that maximizes this objective function. To accomplish this, the discrete search algorithm takes small steps in the sentence space by making word-level editing, e.g., insertion, deletion, and replacement. Note that with search steps defined this way the dimension of search states grows exponentially with the vocabulary size. Hence, an exhaustive search is infeasible. Efficient heuristic search would be the more appropriate algorithm. The search algorithm can be plugged in with simple Hill-climbing or other non-greedy variants such as simulated annealing. This iterative local search process will explore the sentence space within the budgeted time steps, before settling with a locally optimal solution.
	
	Stochastic search formulation of unsupervised text generation has demonstrated its flexibility and capability for a variety of tasks. Applications of the search-based framework in paraphrase generation \cite{li-etal-2018-paraphrase}, document summarization \cite{HC}, and sentence simplification \cite{kumar-etal-2020-iterative} have achieved state-of-the-art performance in respective tasks. Each of these models is equipped with both general and task-specific objective functions to guide the search towards appropriate output for the given task. Since local search algorithms such as hill-climbing (HC) and simulated annealing (SA) generate text by directly modifying the input text, search-based frameworks are especially suitable for tasks with significant overlap between input and output.
	
	Despite the success of the search-based framework in a variety of tasks, there are some drawbacks to be tackled. In the aforementioned search-based framework, quality of the generated text is ensured by the heuristically defined objective function, which could provide general guidance on a population level, but might not be specifically instructive for each single sentence. More precisely, the objective functions are believed to correlate with the evaluation metrics on average for a large number of samples, but such correlation is weak when it comes to each individual instance. Hence, the search algorithms would have to face a very non-smooth objective landscape when searching for solution. Detail analysis is presented in the experiment section.
	
	The non-smooth function landscape poses two main challenges for the search algorithm: guidance by the objective function may be inaccurate, meaning that the objective function which the search algorithm aims to maximize is skewed from the true measure of success; search algorithms would be more susceptible to local optimum in a non-smooth function landscape, leading to the search frequently getting trapped.

\section{Thesis Statement and Contributions}
In this dissertation, we address the research problem of smoothing the objective function for word-level local search, in order to improve search-based unsupervised paraphrase generation. More specifically, we claim that:

\emph{In the context of heuristic search-based text generation, learning of the search dynamic can help smoothing the objective function in a bootstrapping fashion, thus improving the performance of text generation in terms of BLEU and iBLEU evaluation metrics.}

To attain our objective, we make use of deep neural networks to learn and model the search dynamic. Firstly, the search algorithm would search in the sentence space in order to generate search trajectory samples. Then, we train our deep neural network-based models using the collected search trajectory samples. Finally, the learned models will be used to guide the search algorithm together with the original objective function for the final generation of text. All of these are done in the unsupervised setting. Qualitative and quantitative experiments and analysis will be performed to support the above statement.

In this dissertation, we propose three approaches for learning the search dynamic of search-based text generation. In particular, we employ the Transformer architecture as the backbone for all three of our models. For the final inference, the learned model will be combined with the original objective function to guide the search.

In summary, the main contributions of this dissertation are as follows:
\begin{enumerate}
    \item We reproduce a search-based text generation framework, UPSA \cite{liu-etal-2020-unsupervised}, and validate its performance by evaluation metrics, serving as a testbed for our proposed models.
    
    \item We collect, clean, and process a dataset of search trajectories to train our models.
    
    \item We propose and implement three different novel learning-for-search text generation models. We evaluate and analyze the model quantitatively and qualitatively on the task of paraphrase generation, showing the effectiveness of our proposed approaches.
\end{enumerate}

\section{Thesis Organization}
The thesis is organized as follows: in Chapter 2, we overview the current state of the art of natural language processing and natural language generation. In particular, we review how various kinds of deep neural networks are used to process, represent, and generate natural language. We further explain the general working mechanism of the Transformer backbone that we adopted. Search-based text generation framework is the focus of this dissertation, thus recent work in this area would also be covered in details in this chapter.

In Chapter 3 we detail our reproduction of the UPSA \cite{liu-etal-2020-unsupervised} model as our testbed, and propose three different models stemming from the same idea: two of them are BERT-based \cite{devlin-etal-2019-bert} regression models that aim to directly model the manually designed heuristic objective function, while the third model learns the search via a Transformer-based sequence-to-sequence model \cite{NIPS2014_a14ac55a}. The learned model would be combined with the original objective function to guide a second iteration of search to generate the final output. The motivation of this approach, and how it works with the UPSA model would be discussed in details. Moreover, we also describe how to collect and process a dataset needed to learn from the search dynamics of simulated annealing on the task of unsupervised paraphrase generation.

Experiment and analysis procedures and results are presented in Chapter 4. The main purpose of this chapter is to demonstrate the effectiveness of our proposed models. Following previous work in paraphrase generation, we adopt BLEU \cite{papineni-etal-2002-bleu} and iBLEU \cite{sun-zhou-2012-joint} as measures of success. We also aim to gain insights on the underlying behavior of our proposed models, and verify our motivation of the approach. Specifically, we are interested in how the score predicted by our models correlate with true measure of success, compared with the original objective function; we would also look into how the acceptance ratio in simulated annealing is changed by our models; and finally, we want to investigate whether our proposed models indeed help the search algorithm avoid getting stuck at local optimum.

Finally, we conclude this thesis in Chapter 5 with a summarization of our finding and contribution. We also discuss our thoughts and understanding in retrospect, hoping to pinpoint future direction for search-based text generation.

%% file: tex/background.tex
\chapter{Background and Related work}
\section{Natural Language Generation}
\begin{figure}[!t]\centering
    \includegraphics[width=10cm]{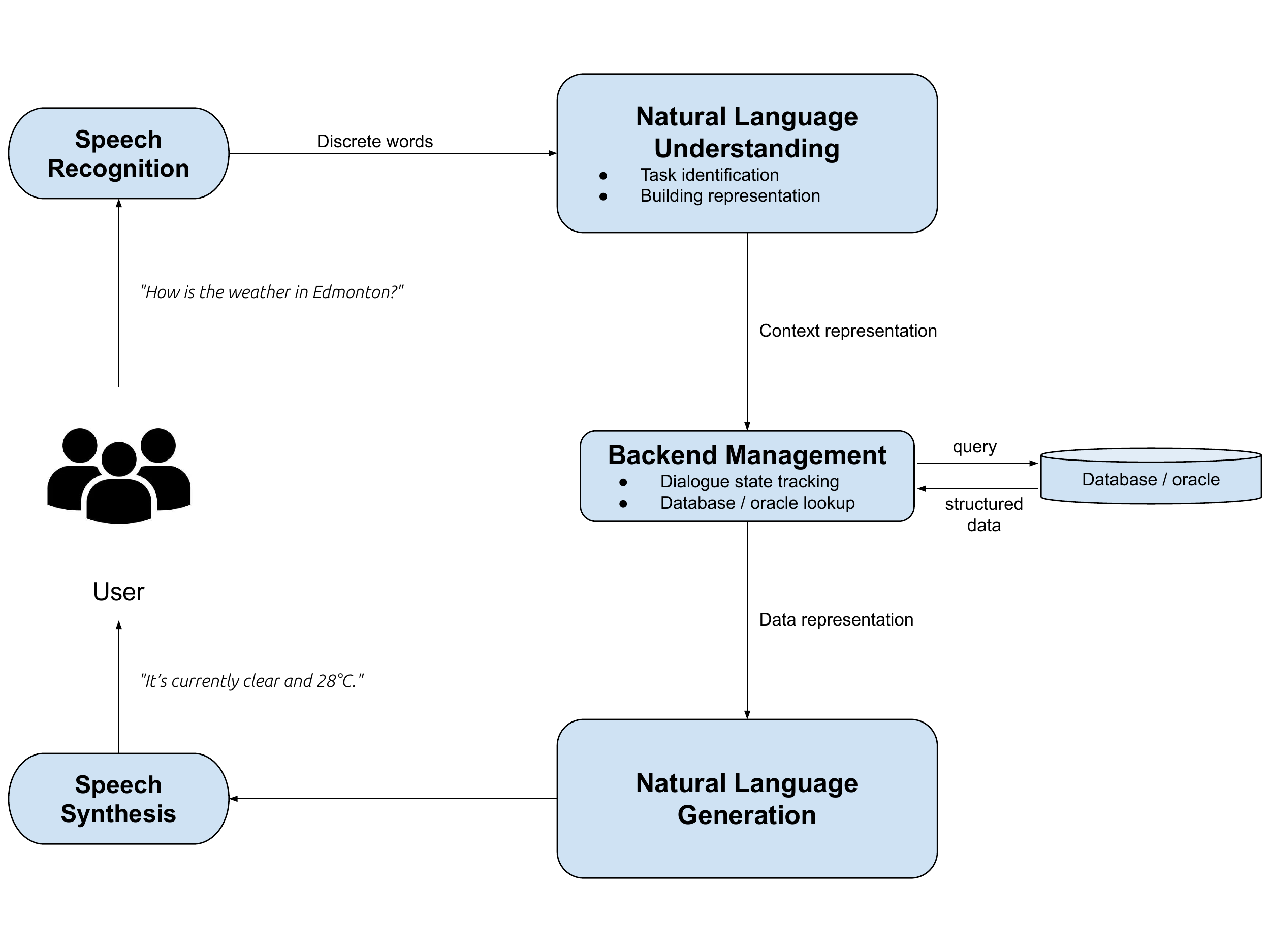}
    \caption{An example dialogue system pipeline.}
    \label{dialogue_system}
\end{figure}

Natural language generation (NLG) has been a long-standing task in the field of natural language processing (NLP). In general, an NLG system aims to generate natural languages that are diverse and able to convey information of interest. NLG and natural language understanding (NLU) are the two important subfields of NLP research. Complementary to NLG, NLU modules are responsible for the understanding of natural language input and representing the semantic and logical relationship with the context. A large number of applications are achieved by the combination of NLG and NLU, yielding NLP systems that can take natural language as input from the user, then process the command or query with a database, and finally respond to the user in natural language as in Figure \ref{dialogue_system}. Although usually taking up different roles in an NLP pipelines, NLU and NLG are by no means orthogonal: in order to generate natural language text, NLG systems need to develop an understanding of natural language. During the text generation process, the NLG system needs to maintain a representation of the semantic meaning or message to be delivered, then generate a text based on that representation. Hence, NLU can be considered as an important building blocks for NLG systems.

\subsection{Text Generation Applications}
Various applications have been realized by NLG models. Perhaps the most sought after application of NLG is machine translation (MT) \cite{bahdanau2014neural}\cite{luong-etal-2015-effective}\cite{wu2016google}, whose goal is to generate text in the target language domain with same meaning as the input from the source language domain. Machine translation is not a trivial task as word-to-word substitution does not solve the problem due to the different grammar and expression style among various languages. To accomplish the task of automatic translation the system needs to understand the syntax and semantic of both the source language and target language, while developing a mapping between the two languages. 
Another commercially successful applications of NLG in recent times is data-to-text generation \cite{wiseman-etal-2017-challenges}\cite{puduppully2019data}. Generating weather report in natural language \cite{goldberg1994using} is one of the earliest application of data-to-text generation, whilst nowadays numerous applications are made to interpret and summarize financial and business data \cite{la-quatra-cagliero-2020-end}\cite{el-haj-etal-2020-financial}.
Another application with rising popularity is dialogue systems, whose goal is to converse with human in natural language, and be able to answer the queries or execute the commands according to the interaction history with the user. In this case, the NLG component would be crucial to process the retrieved data and able to deliver the gist to the user via data-to-text generation. For example, if a user requests ``Summarize the news for me today.'', the NLU component would need to understand the information to look up, and deliver the retrieved data to the NLG component, before the NLG component can perform the summarization task on the news. Oftentimes, the generated natural language response might require further polishing before sending out to user. For instance, the text generated by the aforementioned pipelines might require another model to paraphrase it to improve fluency and correct the grammar, or some sensitive words need to be filtered out.

Challenges of both NLU and NLG are rooted in the complex nature of human languages: compared with programming or scripted languages that come with rigid structures and unambiguous instruction, human languages are more likely to be ambiguous, diverse, context dependent, and even oftentimes syntactically erroneous. For this reason, natural languages with appropriate syntax and semantic patterns are in general difficult to specify by rules and templates. More specifically, the challenge of NLU is to produce a consistent representation for the same semantic meaning in various forms, while the challenge of NLG is to stay consistent throughout the process of emitting words without degeneration. Another major challenge for NLG is the vocabulary size. Take English language as an example. Typically, the vocabulary size is over 30,000. Considering the number of combinatorial sentences grows exponentially with the length by a factor of the vocabulary size, such high dimensionality renders exact solutions intractable even if the true measure of success is known.

\subsection{Task Formulation}
The general task of generating text can be formulated as follows: Given the source language domain $\mathcal{X}$, the target language domain $\mathcal{Y}$, and a task specific function $f(X,Y)$ indicating the quality of the generated text $Y \in \mathcal{Y}$ given the input $X \in \mathcal{X}$, the goal of text generation would be to find the best text $Y^*$ that maximizes the task specific objective function $f(X,Y)$. For example, in machine translation $f(X,Y)$ would evaluate the combination of semantic similarity between the source text and translated text, fluency of the generated text, as well as how appropriate the tone is; for automatic document summarization, $f(X,Y)$ would be based on the preservation of the original meaning and some measure of brevity. However, such $f(X,Y)$ may not be easily defined due to the difficulty of evaluating complicated natural languages. Ideally this should be done by human evaluation. Oftentimes we might need to settle for an approximate evaluation metric that is largely representative of the conceptual idea, and feasible for automatic evaluation.

\subsection{Modelling Text Generation}
Traditionally, text generation is accomplished by rules and templates. For relatively narrow applications, filling the blank in a fixed template would be viable. For example, weather report is one of the more narrow domains where everyday almost the same data structure is fed to the text generation system to generate the weather report, meaning that it would suffice to have one or a few templates with the same blanks to be filled in by the fixed data structure (e.g., temperature and humidity). To handle slightly varying data structures, the filling-the-blank system can be embedded by scripting languages, in which one can specify rules that constitute the templates using conditional branching, or logical loops to handle slightly varying data structure. However, as seen such system is manually constructed by humans; thus, expressions are relatively restrictive because the system would not be able to generate text that was not directly scripted by humans. Text generated in such a way is by no means diverse, and is considered not to be a form of natural language. As seen, although using templates and rules might work well in some narrow applications, such system would obviously fail when facing more complicated tasks that cannot be scripted for every instances, not to mention even if the scripts are achievable by intensive labor, they are impossible to generalized to different domains.

In recent years, the strong function approximation capability of deep neural networks (DNN) \cite{goodfellow2016deep} has revolutionized the field of NLP. DNNs are complex composition of linear functions with non-linear composite connections that can theoretically approximate any function given enough memory and computational capacity \cite{hornik1989multilayer}. Multi-layer perceptron is a common class of DNN architecture consisting of multiple layers of linear neurons that are parameterized by weights and bias, with non-linear synapses connecting layers to layers in one direction. Typically DNNs operate as feed-forward connectionist networks that take the input and let the numerical data flow through layers by layers, producing representations from each layer. Then the representation can be used by downstream decoders to generate text. Such design is originally inspired by how synaptic system in human brains transmit signals. 

One key advantage of DNNs over traditional rules and templates is the elimination of manual feature engineering. DNNs are able to learn a representation of content automatically from data and use it to make prediction. Tremendous successes have shown the capability of DNNs learning meaningful features and using them to model noisy, uncertain, and high-dimensional environments.

The power of the DNNs comes at the price of needing huge amount of data for training. Typically DNNs are trained by trial and error. At each training step, the DNN makes a prediction, which is evaluated by a differentiable loss function to indicate how good the prediction is. Then, automatic learning of DNNs is accomplished by backpropagation (BP), which is the process of modifying the weight and bias parameters in all neurons in the direction of negative gradient with respect to the loss function by a small step. The intuition of modifying parameters in such a way is that weights moved in the direction of negative gradient lead to steepest descent of the loss function locally. By making small gradual local steps for a great many iterations, the loss function should be somehow minimized.
\begin{figure}[!t]\centering
    \includegraphics[width=10cm]{\main/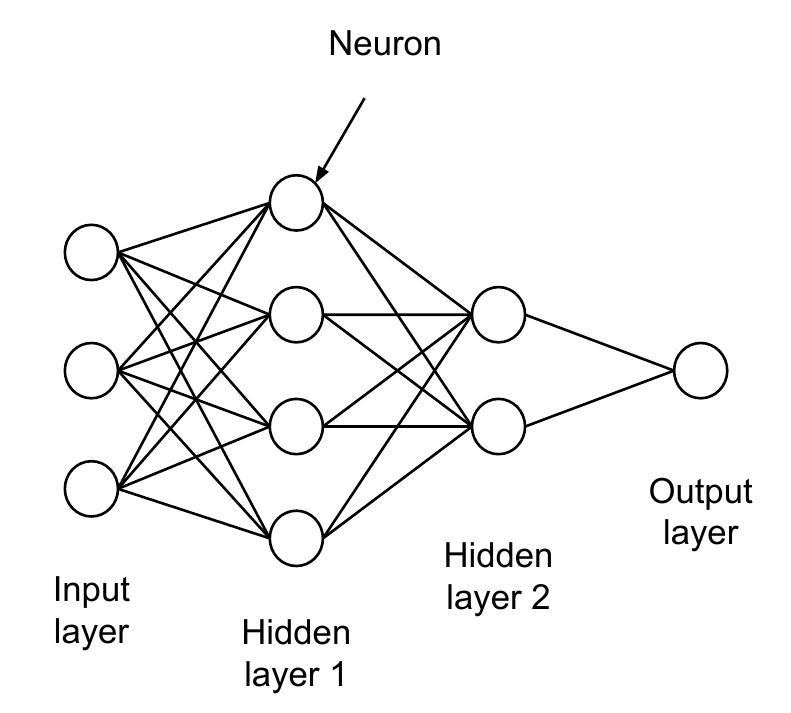}
    \caption{An example deep neural network.}
    \label{dnn}
\end{figure}

Learning and predicting the probability $P(w_1, \dots, w_l)$ of a sequence ${w_1, \dots, w_l}$ of length $l$ are the backbones of various NLP applications, including speech recognition, part-of-speech tagging, and semantic parsing. Since text is made of discrete word tokens, such probability is usually factorized into $P(w_1, \dots ,w_l) = P(w_1) \cdot P(w_2|w_1)\cdot \dots \cdot P(w_l|w_1,\dots,w_{l-1})$. In other words, the probability of a given sentence is computed by the product of probability of all words given all preceding words. Before the era of DNNs, lack of computational power calls for the need of a simplified approximation. Uni-gram and $n$-gram models are designed for this exact purpose, which essentially approximate $P(w_1, \dots ,w_l) = P(w_1) \cdot P(w_2|w_1)\cdot \dots \cdot P(w_l|w_1,\dots,w_{l-1})$ with truncated probabilities $P_\text{uni-gram}(w_1, \dots ,w_l) = P(w_1)\cdot P(w_2) \cdot \dots \cdot P(w_l)$ and $P_{n\text{-gram}}(w_1, \dots ,w_l) = \prod_{i=1}^{l} P(w_i | w_{i-(n-1)}, \dots, w_{i-1})$, respectively. Modern language models parameterized by DNNs are more expressive, and thus can fully model $P(w_1, \dots ,w_l) = P(w_1) \cdot P(w_2|w_1) \cdot \dots \cdot P(w_l|w_1, \dots, w_{l-1})$. In order to convert word tokens into numerical representation for DNNs, words are represented by word embeddings, which map discrete word tokens to continuous real-valued vectors while encoding semantic meaning of words in such a way that words with more similar meaning are mapped to vectors with a smaller distance, and vice versa. Representing words in such way provides more meaning than simply using indexes. Moreover, modelling a vocabulary of words in continuous space theoretically and practically alleviates the curse of dimensionality. Plenty of pre-trained embeddings are readily available nowadays, such as Word2vec \cite{word2vec} and GloVe \cite{pennington2014glove}, setting up the foundation for various applications.

\subsection{Sequential Models}
\begin{figure}[!t]\centering
    \includegraphics[width=10cm]{\main/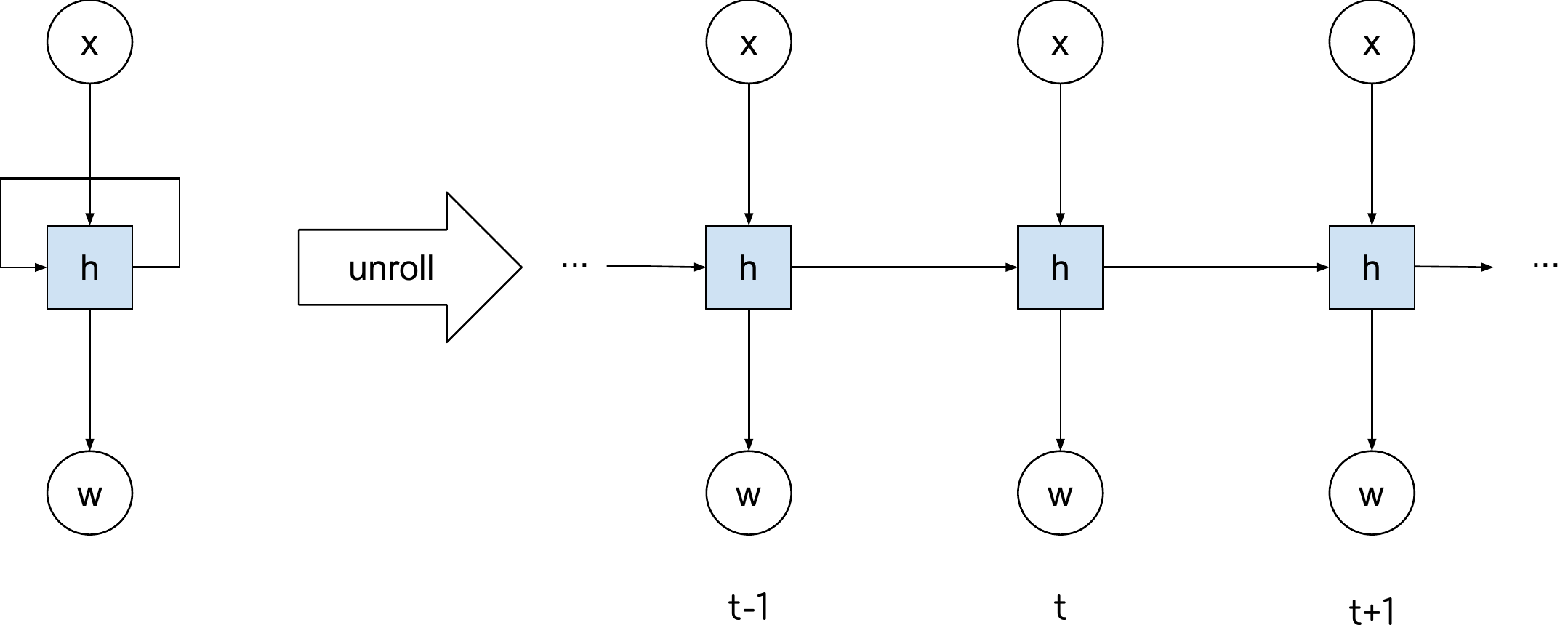}
    \caption{Recurrent neural network and unrolled perspective.}
    \label{rnn}
\end{figure}

Recurrent neural network (RNN) is a family of DNN models that connects computational layers in a cyclic way over temporal steps to process varying-length sequences sequentially and finally outputs a representation for downstream tasks. As for vanilla RNNs, the outputs of the neurons at time step $t$ are used as input for the neurons in time step $t+1$. In such a way, each time step the representation is being processed using the same set of weights. Recurrent computation can be conceptually unrolled to reflect the feed-forward pass over time as in Figure \ref{rnn}. Theoretically, vanilla RNNs should have the power of modelling long-term dependencies. However, in practice vanilla RNNs face various problems, including gradient exploding and vanishing during training time \cite{hochreiter1998vanishing}\cite{jozefowicz2015empirical}. Long-short term memory (LSTM) \cite{hochreiter1997long} recurrent neural networks alleviate these issues by regulating dependencies using a gating mechanism.

As seen, vanilla RNN is one-directional, meaning that when processing word $w_t$, only ${w_1, \dots, w_{t-1}}$ are taken into consideration. Bidirectional RNNs is a simple direct improvement over vanilla RNNs. Instead of only having one hidden state accumulating information from the beginning of a sequence, bidirectional RNNs use another hidden state that accumulates information from the end of the sequence, providing more context when dealing with each single word in the sequence.

The sequence-to-sequence (Seq2seq) \cite{NIPS2014_a14ac55a} model is a popular text generation model for various tasks, which essentially maps an arbitrary length text sequence to another arbitrary length sequence by compressing the former into a latent representation before generating output conditional on that. The Seq2seq model is originally developed for the task of machine translation due to its flexibility with different source and target language domains, but its simplistic and flexible design has made its way to numerous applications, such as part-of-speech (POS) tagging. Seq2seq models are typically constituted by an encoder, which can be parameterized by RNNs, LSTMs, or other sequential models, and a decoder that generates output sequentially. The encoder takes an input sentence, and outputs a latent representation that carries the information from the input sentence. The decoder then takes the latent representation as input and generate text auto-regressively from the first word until a special stopping token is emitted, to deliver the content specified by the latent representation. At each time step, a probability distribution over the entire vocabulary conditional on the already generated tokens $P(w_t|w_1, \dots, w_{t-1})$ is given by the softmax function in the decoder, indicating the probability of $w_t$ following a sequence of $w_1, \dots, w_{t-1}$, given the input sentence.
\begin{figure}[!t]\centering
    \includegraphics[width=10cm]{\main/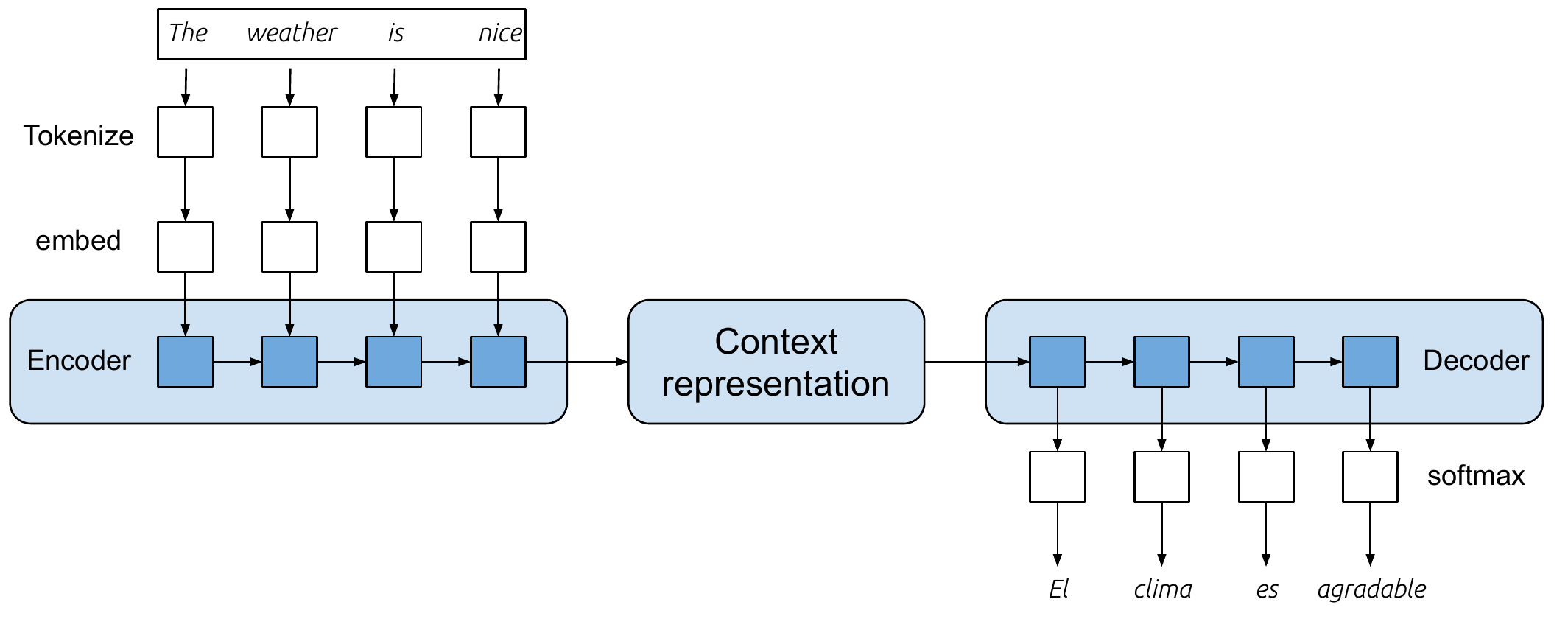}
    \caption{An example sequence-to-sequence (Seq2seq) model for English-to-Spanish machine translation.}
    \label{seq2seq}
\end{figure}

\subsection{Attention Mechanism}
Despite the effectiveness of the gating mechanism in LSTMs, RNN-based Seq2seq models still suffer from two major issues: RNNs are not parallelizable since the input for a time step strictly depends on the previous time steps, thus computation in all time steps must be done one-by-one; although the gating mechanism in LSTMs can alleviate the gradient vanishing problem, the gradient exploding problem implies the difficulty of feeding hidden state linearly forward over time steps (e.g., if $w_{t+n}$ is heavily dependent on $w_t$, then the hidden state needs to carry the relevant information from time step $t$ all the way through time steps $t$ to $t+n$). The attention mechanism \cite{bahdanau2014neural}, inspired by cognitive science, is designed to tackle these issues. Instead of using the hidden state from strictly the previous time step (forward or backward), an attention mechanism also makes use of the input text directly weighted by their relevance to the output. The attention weights given by an alignment model parameterized by DNN is a probability distribution indicating the ``correlation''  between all the words in input sentence and the specific word being processed.

Transformers \cite{transformer} are a family of sequential model that uses attention only to process text and has been trending since it was first introduced. Similar to the aforementioned RNN-based Seq2seq model, Transformer-based Seq2seq models also follow the encoder-decoder architecture. Other than that, Transformers are very different from RNNs in terms of architecture and have several advantages. The core idea of Transformers is to use attention alone without recurrent structure to process sequential data to allow the model to process time dependencies without linear recurrent passage. Besides, the attention mechanism enables parallelization of computation, and thus improving training efficiency. The attention mechanism used by the original Transformer \cite{transformer} is multi-head self-attention, meaning that when processing the input the model attends to the input itself, and having multiple attention heads functioning independently.

Various large-scale Transformer-based models have demonstrated their capability in a wide variety of tasks. Bidirectional Encoder Representation from Transformer (BERT) \cite{devlin-etal-2019-bert} is a set of models for NLP tasks developed by Google. The original English-language BERT models are pre-trained on the Bookcorpus \cite{zhu2015aligning}\footnote{Bookcorpus dataset: \url{https://github.com/soskek/bookcorpus}} and the English Wikipedia dataset\footnote{English Wikipedia dataset: \url{https://en.wikipedia.org/wiki/Wikipedia:Database_download}}, consisting of 800M and 2.5B samples, respectively. BERT achieved state-of-the-art performance on a variety of NLU benchmarks, including General Language Understanding Evaluation (GLUE)\footnote{GLEU:\url{https://gluebenchmark.com/}}, Standard Question Answering Dataset (SQuAD)\footnote{SQuAD: \url{https://rajpurkar.github.io/SQuAD-explorer/}}, and Situations with Adversarial Generations (SWAG) \cite{zellers2018swag}, showing the superior capability of BERT in language understanding. Representations learned by BERT while training towards general purpose NLU tasks are now used as embeddings for various downstream tasks. Evidence \cite{rogers2020primer} shows that pre-trained BERT models often performs well with only minimal fine-tuning on relevant NLU tasks. Generating text from BERT is not trivial, as BERT is inherently bidirectional, rendering auto-regressive word generation infeasible. However, there are various ways to decode text from BERT, such as starting with a sentence filled with empty blanks, then using BERT to fill all the blanks in a specified order \cite{havens2019fitbert}. Generative Pre-trained Transformer (GPT) \cite{gpt} and its direct scaled up successor GPT-2 \cite{gpt2} are pre-trained Transformers geared more towards text generation. Different from BERT, GPT models have only decoder, but are also equipped with multi-head self-attention.

\section{Edit-Based Text Generation}
Different from the MT-based Seq2seq model that emits words one after another, edit-based text generation models refer to the family of models that learn to generate text by modifying the input text. Solving NLG tasks by predicting edit operation sequence has several advantage over auto-regressive text generators: many monolingual text generation tasks require generating text that have big overlap with the input. Take sentence simplification as an example: the easy way to simplify a sentence is to remove non-essential utterances while keeping all important key words, then fill in some new words to make the simplified sentence fluent and grammatically coherent. Using a MT-based Seq2seq model for this purpose would require the model to implicitly learn all these edit actions, while still be able to recover the essential keywords via the context representation. Indeed, experiment results show that Seq2seq models in this case are prone to generate the exact same output as the input most of the time, indicating that the minimal differences between inputs and outputs are oftentimes not captured by the Seq2seq models \cite{zhao-etal-2018-integrating}.
\begin{figure}[!t]\centering
    \includegraphics[width=10cm]{\main/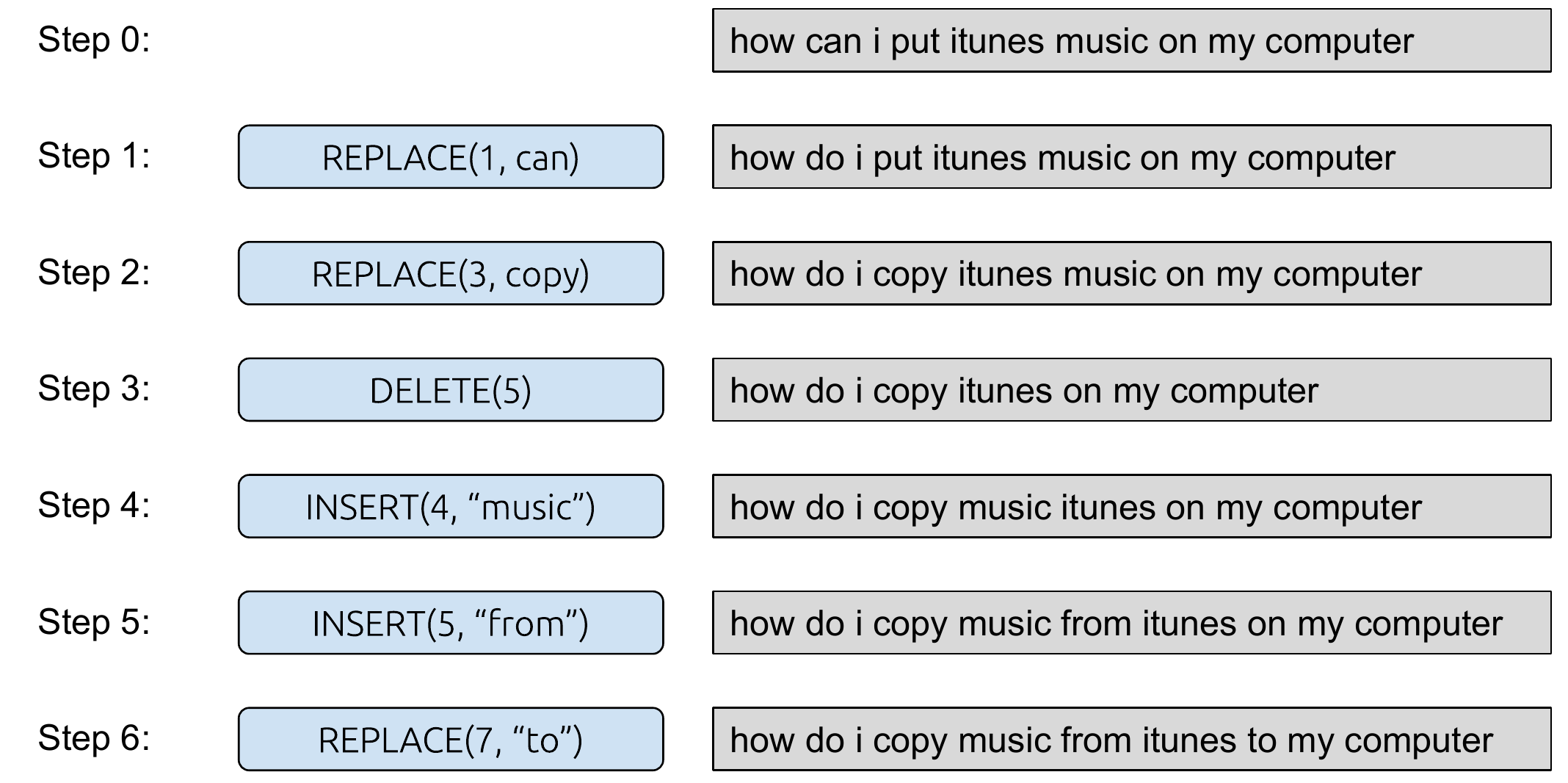}
    \caption{Example of paraphrase generation by edit.}
    \label{edit}
\end{figure}

Learning the edit operations in the supervised setting requires first finding the edit operation sequence given parallel sentence pairs. This can be done by dynamic programming (DP) by first creating a table to store the results of sub-problems(i.e. partially edited text), then the minimum edit operation sequence can be obtained by tracking the table. Each valid edit label consists of a sequence of edit operation that if executed will convert the input sentence $X$ to the target sentence $Y$. Commonly used word-level edit operations are insertion, deletion, and replacement. However, various operation formulations can be used to suit the need of various tasks. Edit-based approach allows the model to theoretically generate any sentence while also be able to preserve important keywords from the input sentence relatively easy by choosing not to modify those tokens. Moreover, such design also comes with more controllability and interpretability than MT-based Seq2seq models, as the generated operation sequence can reflect which part of the input needs to be edited instead of generating everything from scratch as a black-box.

\subsection{Supervised Learning}
In this section, we overview the current state of the art of edit-based text generation in supervised setting.

\cite{alva-manchego-etal-2017-learning} aligns the input and output sentence to the best extent, before using heuristics to find the edit sequence label. The word-level edit operations they adopt are $\textsc{MOVE}$, $\textsc{DELETE}$, and $\textsc{REPLACE}$. Since heuristics are used instead of dynamic programming, annotations themselves are not exactly optimal, yielding $92\%$ accuracy and an average of $0.7$ F1 score compared with human expert annotations. Finally, they train a bi-directional LSTM-based RNN model to learn from the automatic annotated edit labels. Experiment results on sentence simplification task show that the RNN model does not learn the edit operations very well, but is still able to generate good quality sentence in terms of human evaluation.

EditNTS \cite{dong-etal-2019-editnts} adopts the neural programmer-interpreter model for sequential text edit operation prediction. The programmer module predicts the edit operation by jointly considering the partially generated sentence, edit trajectory, and contexts, while the interpreter is in charge of the realization of edit operations. Different from the RNN model \cite{alva-manchego-etal-2017-learning}, edit annotations in EditNTS are labeled by dynamic programming, which theoretically should yield 100\% accuracy, superior to the heuristic based approach. For the task of sentence simplification, the programmer predicts one operation from $\textsc{ADD}$, $\textsc{Delete}$, and $\textsc{KEEP}$ for each word in the sentence. EditNTS model is trained to predict the correct edit label for each word by maximum likelihood training. The maximum likelihood training loss function can be modified to counter the class imbalance among $\textsc{ADD}$, $\textsc{Delete}$, and $\textsc{KEEP}$ operations by assigning different weight to each edit label in the loss function. This is also one way to inject human expert bias of preference among edit operations (i.e. to edit more or to keep more words from the input sentence). Experiment results on sentence simplification task with WikiLarge, WikiSmall \cite{wiki_dataset}, and Newsela dataset \footnote{Newsela dataset: \url{https://newsela.com/data/}} show that on all three dataset EditNTS consistently outperforms all other simplification models that does not use external knowledge base by the time of publication. More specifically, performance of EditNTS is consistently higher than that of the vanilla MT-based RNN model \cite{alva-manchego-etal-2017-learning}, demonstrating the superiority of edit-based text generation in sentence simplification task. We attribute the higher performance of EditNTS to two factors: quality of edit labels used by EditNTS are of higher quality since dynamic programming are used to create the labels; the programmer-interpreter model is more suitable than simple RNNs for learning edit labels.

Levenshtein Transformer \cite{NEURIPS2019_675f9820} is an interesting mixture of edit-based model and auto-regressive model. The editing process is done by first processing input by a Transformer model, then iteratively delete, insert placeholder, and finally fill in the placeholder by word. The deletion, placeholder insertion, and fill-in-blank steps are each done by their own models at each step, but share the same Transformer backbone. Such iteration loops until the deletion and insertion policies stabilize or fixed budget is reached. In the experiment, Levenshtein Transformer achieves high performance and efficiency in machine translation, text summarization, and automatic post-editing.

LaserTagger \cite{malmi-etal-2019-encode} proposed an edit-based text generation model using pre-built phrase vocabulary. This model comes with a very simplistic three step design: encode input, tag edit operations, and realize the edits. The highlight of this work is the phrase vocabulary they adopt for phrase insertion. The phrases in the vocabulary are built from pair-wise alignment of all parallel pairs in a dataset, then the mismatched $n$-grams between input and ground-truth output after alignment are stored in the phrase vocabulary. The phrase vocabulary has a much smaller size than a word vocabulary, and thus making the model eaiser to train. Using such a restricted vocabulary, LaserTagger yields similar performance compared with the BERT-based Seq2seq model when large number of training examples are available, but significantly outperforms the Seq2seq model when training examples are scarce ($\leq 1000$).

\subsection{Unsupervised Text Generation}
As seen, edit-based text generation models have several advantage over MT-based  models in supervised learning setting. However, lack of parallel corpora in certain tasks call for unsupervised edit-based models. 

The variational auto-encoder (VAE) \cite{kingma2013auto} is a popular generative model for unsupervised setting. \cite{bowman-etal-2016-generating} proposes to generate text using VAE by sampling from the latent sentence space. Different from auto-encoders (AE) \cite{ballard1987modular}, VAE compresses the input (i.e. a sentence) into a mean $\mu$ and standard deviation $\sigma$ that parameterize a multi-variate distribution, which is then used to sample the latent representation $z$. Then the LSTM-based decoder would generate text conditional on the latent representation. The training objective is to minimize the evidence lower bound, which measures the distance between the estimated posterior and the real posterior. Experiment results show that VAE is a competitive approach for imputing missing words and text classification. However, due to the black-box nature of the latent space, VAE would not be able to generate text with specific attributes.

Generating text by Metropolis-Hastings (MH) sampling in word space \cite{Miao_Zhou_Mou_Yan_Li_2019} is the cornerstone that leads to search-based NLG which this dissertation focuses on. Metropolis-Hastings algorithm is a Markov Chain Monte Carlo (MCMC) sampling algorithm that generates samples by iteratively jumping from state to state within a Markov Chain governed by a discrete probability distribution. Here the Markov Chain is defined to be the space of all sentence $\mathcal{X}$, and each state is a sentence $x \in \mathcal{X}$. To explore the sentence space by sampling, the MH algorithm involves two steps: starting from any state $x$, first a candidate transition to $x'$ is proposed using one of the word-level edit operations from insertion, replacement, and deletion; Then, the probability of accepting this proposed transition $x'$ is determined by 
\begin{equation}
    A(x'|x) = \operatorname{min}\{1,A^{*}(x'|x)\}
\end{equation}
\begin{equation}
    \text{where} \quad A^{*}(x'|x) = \frac{\pi(x')g(x|x')}{\pi(x)g(x'|x)}
\end{equation}
where $g(x'|x)$ is the \emph{proposal function} that suggests a tentative transition $x'$ from $x$, and $\pi(x'|x)$ is the \emph{stationary distribution} that defines a probability distribution over the sentence space. As seen $A(x'|x)$ is rectified to the range of $[0,1]$ to output a probability of accepting the transition to $x'$. In other words, the probability of not accepting transition to $x'$ and remaining at $x$ is $1-A(x'|x)$. The motivation of MH algorithm is that sampling directly from $\pi(\cdot)$ is difficult as it involves computing $\pi(x)$ for all $x$ to get a normalized measure. In the MH algorithm, only state at the sampling time $x_t$ and the proposed neighbour $x'_{t}$ require inference with $\pi(\cdot)$, thus bypassing the computational complexity of a global inference. For the task of text generation, $\pi(\cdot)$ is essentially a non-learnable heuristic function that takes as input a specific state $x$ (the original input sentence $x_0$ is also taken as input implicitly), then evaluates a score indicating the quality of the state $x$ with respect to the task. The idea of using MH sampling is to generate a sequence of states that follows the distribution $\pi(\cdot)$, such that sentences with relatively higher probability (i.e. higher quality sentence) are likely to be sampled more, while those with relatively low probability (i.e. lower quality sentence) are less likely to be sampled. In other words, MH sampling would be able to continually generate different sentence states by jumping around in the sentence space.

Model in \cite{Miao_Zhou_Mou_Yan_Li_2019} does not require any supervision since the distribution $\pi$ is manually designed by human to reflect the requirement for the text to be generated $x^{*}$. Consider the example of keyword-to-text generation. The distribution $\pi{\cdot}$ can be set to 
\begin{equation}
    \pi(\cdot) \propto p_\text{LM}(\cdot) \cdot s_\text{keyword}(\cdot)
    \label{pi_keywords}
\end{equation}
where $p_\text{LM}(\cdot)$ is the probability given by a language model, indicating the fluency and grammatical coherence of a particular sampled state. More specifically, the language model for evaluating $p_\text{LM}(\cdot)$ is trained from non-parallel corpora in a unsupervised fashion; $s_\text{keyword}$ is the hard keywords constraint taking value of 1 or 0, indicating if all required keywords are present in the state. This ensures the MH sampler never visit any state that does not strictly have all the required keywords present in the sentence.

Furthermore, for the task of paraphrase generation, the MH algorithm would sample from
\begin{equation}
    \pi(\cdot) \propto \pi(\cdot) \propto p_\text{LM}(\cdot) \cdot s_\text{match}(\cdot|x_0)
\end{equation}
where $p_\text{LM}(\cdot)$ is language model probability indicating the fluency of generated text, similar to Equation \ref{pi_keywords}; $s_\text{match}(\cdot|x_0)$ is a matching score function indicating how similar are the semantic meaning between the generate sentence and original sentence to be paraphrase, which could be implemented by cosine similarity of pre-trained embeddings, or skip-thoughts sentence similarity \cite{kiros2015skip}. Note that although manually designed heuristics are used, there are no explicit rules or templates being imposed. Components of $\pi(\cdot)$ (except for hard constraint) are all parameterized by pre-trained DNNs or embeddings, and thus able to recognize complicated semantic structures.

Experiment result in \cite{Miao_Zhou_Mou_Yan_Li_2019} shows that the MH sampler yields promising performance in keywords-to-sentence generation, unsupervised paraphrase generation, and unsupervised error correction, showing the generality of their model. Hence, it can be concluded that it is feasible to manually design a task specific distribution $\pi(\cdot)$, then generate text via sequential sampling from such distribution.
\begin{equation}
    x_1,x_2,\dots,x_T \sim \pi(\cdot|x_0)
    \label{sampling_based}
\end{equation}

MH sampling is in a sense wasteful as it samples both high quality and low quality sentences. Although sentences of higher quality are more likely to be sampled, there are still quite some steps wasted on low quality region in the sentence space as in Equation \ref{sampling_based}. In other words, it would make more sense to use a sampling mechanism that focus on finding a maximum score sentence as in Equation \ref{max_sentence}.
\begin{equation}
    x^{*} = \underset{x}{\operatorname{argmax}} \pi(x|x_0)
    \label{max_sentence}
\end{equation}
In the next section, we overview search-based sampling for this exact purpose.

\section{Search-Based Text Generation}
In this dissertation, we refer search-based text generation models as word-level edit-based models that generate text by searching towards an objective function. Similar to \cite{Miao_Zhou_Mou_Yan_Li_2019}, a manually specified objective function is used to indicate the quality of an arbitrary sentence for a given task. However, instead of sampling from a distribution, search-based model aims to optimize the objective function by searching towards its optimum using discrete word-level edit operations.

\cite{HC} tackles the task of document summarization by hill-climbing(HC) search, which ensures every accepted search step leads to a higher scored state. Summarization in this model is realized by word extraction, meaning that the model generates summary by directly using words from the original document. A good summary should be fluent, semantically similar to the original document, but has a shorter length. To specify these requirement, the objective function designed for the summarization task is
\begin{equation}
    f(\pmb{y};\pmb{x},s) = f_{\overrightarrow{\text{LM}}}(\pmb{y}) \cdot 
    f_\text{SIM}(\pmb{y};\pmb{x})^{\gamma} \cdot 
    f_\text{LEN}(\pmb{y};s)
\end{equation}
where $s$ is the desired length of summary; $f_{\overrightarrow{\text{LM}}}(\pmb{y})$, $f_\text{SIM}(\pmb{y};\pmb{x})$, and $f_\text{LEN}(\pmb{y};s)$ are components enforcing different requirements for generated text; and $\gamma$ is a hyperparameter adjusting the relative weights of these components. $f_{\overrightarrow{\text{LM}}}(\pmb{y})$ is the perplexity of sentence $\pmb{y}$ given by a forward-backward language model, indicating the fluency of sentence $\pmb{y}$; $f_\text{SIM}(\pmb{y};\pmb{x})$ measures the similarity between the original document and sentence $\pmb{y}$, which is realized by computing the cosine distance of the sent2vec \cite{pagliardini-etal-2018-unsupervised} embeddings of the two; and $f_\text{LEN}(\pmb{y};s)$ is the hard constraint of length that outputs 1 if length sentence $\pmb{y}$ is equal to $s$ and 0 otherwise. The search space is defined to be a vector of Boolean variables $(a_1,\dots, a_n)$ indicating if each specific word in the original document is extracted for the summary. To find a satisfactory summary in this search space, the model would iteratively sample one word at a time from the original document to swap a random word from the summary. The HC algorithm would only accept the new summary only if it is evaluated to have a higher score $f(\pmb{y};\pmb{x},s)$ than the current summary. This ensures that as the sampling process goes on, the quality of summary can only get better or stay the same. Experiment results on Gigaword and DUC2004 dataset show that such approach achieves a new stat-of-the-art performance on headline generation at the time of publication.

\cite{kumar-etal-2020-iterative} uses an iterative search algorithm similar to hill-climbing for the task of text simplification. Components of the objective function in this model are as follows:
\begin{equation}
    f(s) = f_{eslor}(s)^{\alpha} \cdot f_{fre}(s)^{\beta} \cdot (1 / f_{len}(s))^{\gamma} \cdot f_{entity}(s)^{\delta} \cdot f_{cos}(s)
\end{equation}
where $f_{eslor}(s)$ is the syntactic log-odds ratio (SLOR) \cite{kann-etal-2018-sentence} given by a syntax-aware language model, which measures fluency and structural simplicity of a simplification. SLOR differs from the plain forward-backward language models used in \cite{Miao_Zhou_Mou_Yan_Li_2019}\cite{liu-etal-2020-unsupervised} in the sense that it penalizes the plain language model probability by length and the product of uni-gram probability of all words in the sentence. This is to factor out the low language model probability of rare words, as they carry important information but oftentimes lead to low language model probability if present in a sentence. SLOR also takes as input the part-of-speech (POS) and dependency tags, claiming better evaluation of language fluency than a plain language model. $f_{fre}(s)$ stands for the Flesch Reading Ease \cite{kincaid1975derivation} score, which measures the readability of a sentence. The inverse of sentence length $(1 /f_{len}(s))$ is designed to give higher scores for shorter sentences. $f_{entity}(s)$ is the count of all named entities, compensating their low language model probabilities. $f_{cos}(s)$ is the cosine distance between the sentence embeddings of the original sentence and the simplified sentence. The edit operations used in this model are both in word and phrase level: removal, extraction, reordering and substitution. At each iteration, each of the edit operation is used to generate multiple candidates, and only those candidates with higher objective scores than the previous step by a threshold would be accepted. Hence, the search algorithm in this model is similar to hill climbing, but have multiple candidate branching at each step. However, the intrinsic issue with only accepting higher score would make the search prone to local optimum. In order to jump out of local optimum, some randomness would be needed to occasionally accept lower score.

UPSA \cite{liu-etal-2020-unsupervised} uses simulated annealing (SA) search algorithm for the task of unsupervised paraphrase generation. SA algorithm differs from the hill climbing search in the sense that non-greedy search steps are occasionally allowed. At each iteration one candidate would be generated from one of the word-level operations: insertion, replacement, deletion. If the new candidate state is evaluated to have a higher score than the current state, the current state is always updated to the candidate; if the new candidate state has a lower score than the current state, SA algorithm would still be possible to accept it with a low probability. Such non-greedy search steps are more likely to be allowed in the beginning of search, and become gradually less likely as the search goes on to settle for a optimum point. 

\subsection{Learning from Search}
Despite the state-of-the-art performance in various tasks by the search-based edit models, there are opportunities for enhancement. One major drawback of the search-based framework is that the search process is oftentimes noisy. This is to be expected since the objective functions are manually designed and not learnable. 

\cite{NEURIPS2020_7a677bb4} proposes an iterative search and learn framework for text generation. The core idea is using a Seq2seq model to provide new starting points for the SA search. In the first learning stage, SA algorithm is first performed to search towards the objective function to generate a dataset with pseudo-labels. The generated pseudo labels are used to train a Seq2seq model by cross-entropy loss. In the second learning stage, beam search is performed on the Seq2seq model to obtain a set of candidate outputs. These outputs are used as starting point for SA search again to generate another set of outputs. The Seq2seq is then trained by max-margin loss to maximize the margin between the highest scored instance from these two sets of outputs and the rest. Such alternation between searching (SA) and learning (Seq2seq) is carried on for a couple iterations. Experimental results show that such iterative alternation of searching and learning indeed improve the performance in paraphrase generation task as more iterations are performed, and finally achieve state-of-the-art performance. They attribute the success of this simple approach to the Seq2seq model effectively smoothing out the noise in search results. Moreover, the positive outcome of such framework shows that learning of search results is feasible. However, learning and searching procedures in this framework both function as black-boxes of input-output correspondence. The natural question along this line would be: Is it also possible to learn from the internal search dynamic?

Attempts are made to use learnable function to improve local search by learning directly from the search dynamics in \cite{rl_value_function}. This framework learns a function that predicts the outcome of a local search algorithm. In other words, a model predicts the expected maximum objective score seen on a trajectory that starts from a given state $x$ and follows the local search algorithm. To sample from the search dynamics, the search algorithm is first performed to collect search trajectories. Then the learnable evaluation function is trained from the collected trajectories to predict the search outcomes. The training framework consists of two stages running alternately: a search algorithm to maximizes the original objective function; and another search algorithm maximizes the learned evaluation function. Output of either stage would be used as input for the next stage for several iterations. Hence, this framework can be viewed as a smart restarting mechanism for the search algorithm. Experimental results on combinatorial games such as bin-packing, rerouting, and Boolean satisability show promising performance. However, this formulation has yet to demonstrate its capability when it comes to notoriously large search space, such as text generation.

%% file: tex/Methodology.tex
\chapter{Methodology}
In Section \ref{section: upsa}, we overview the framework of search-based text generation. Specifically, we focus on the paraphrase generation framework of UPSA, which shows state-of-the-art performance in paraphrase generation according to \cite{liu-etal-2020-unsupervised} and will serve as our baseline for setting up the foundation of our study. We follow \cite{liu-etal-2020-unsupervised} and evaluate our models on the task of unsupervised paraphrase generation.

In Section \ref{section: proposed}, we introduce our proposed models for smoothing the objective function. In particular, we pinpoint the drawback of the current search-based framework, then propose three models for guiding the search process by smoothing the objective function. All three models learn from search trajectories of the SA algorithm. 


\section{Unsupervised Text Generation by Simulated Annealing}
\label{section: upsa}
Search-based approaches have demonstrated their capability in various unsupervised text generation tasks and have been gaining popularity over recent years \cite{HC}\cite{kumar-etal-2020-iterative}\cite{liu-etal-2020-unsupervised}. The search-based framework consists of a manually designed objective function to evaluate the search states with respect to the specified tasks, and a local search algorithm to find the optimum of the objective function in the sentence space. For the particular task of paraphrase generation, the objective function would need to evaluate grammatical fluency, semantic preservation, and expression diversity. The search algorithm would then perform local search in the sentence space by making word-level edit to search states (i.e. sentence), in order to maximize the objective function. Hence, the task of text generation is formulated as an optimization problem. Details will be provided in the rest of this section to overview components in the objective function to enforce the requirements of paraphrase generation.



\subsection{Unsupervised Objective Function}
The objective function plays the important role of guiding the search algorithm by evaluating the objective score of any given search step. Specifically, the objective function can be considered as a mapping that takes as input an arbitrary sentence $\mathrm{x}$ and the context (e.g. original input sentence $\mathrm{x}_0$), and outputs a numerical score indicating the quality of generated sentence $\mathrm{x}$ for the given task. In other words, a state with a high score indicates it satisfies most of the requirement for the task.

Language fluency and syntactical coherence are among the most fundamental necessities of language generation. In the neural network regime, fluency and syntax are measured approximately by the probability predicted by a language model. This probability indicates the relative likelihood of a particular sentence is drawn from a specific dataset. More specifically, the language model itself is a probability distribution over the sentence space. The language model is trained to maximize the log-likelihood of each sentence in the dataset. Hence a high probability assigned to a sentence means it was likely drawn from the same distribution that induces the dataset. Since all sentences in the dataset are presumed to be grammatically fluent and syntactically coherent, a high language model probability correlates with higher language fluency and syntactical coherence:
\begin{equation}
     f_{\text{flu}}(\mathrm{x}) = \prod\nolimits_{k} P_{{\overrightarrow{\text{LM}}}}(w_{k}|w_{1}, \cdots w_{k-1})
\end{equation}
where $k$ is the length of sentence $\mathbf{x}$. The language model used in our experiment is a two-layer LSTM model with 300 hidden units. No parallel supervision is used for training the language model.

The main goal of paraphrase generation is to generate text that is distinct from the original input. Hence, sentence with more different wording compared with the original input should be assigned a higher score. The BLEU \cite{papineni-etal-2002-bleu} score is an automatic measure for lexical similarity, which computes the length-penalized $n$-gram overlaps between two text sequences. Then $1-\text{BLEU}$ would yield higher score for output dissimilar to the input. In this experiment, the lexical dissimilarity of a sentence is measured by $1-\text{BLEU}$ against the original sentence:
\begin{equation}
f_{\text{lex}}(\mathrm{x}|\mathrm{x_0})=(1-\operatorname{BLEU}(\mathrm{x},\mathrm{x_0}))^S
\end{equation}
where $S$ is a hyperparameter controlling the multiplicative weight of the lexical diversity component.

Although different wording is desired, the semantic meaning of the sentence should not change during the paraphrasing process. To enforce this requirement, word-level or sentence-level embedding distance are used to measure the semantic similarity between input and output. In such a way, sentences with similar meaning would have a shorter embedding distance, and vice versa. Such distance metric is not particularly sensitive to any particular word in the sentence as all words are mapped to embeddings. UPSA adopts two ways for measuring semantic similarity: mini-max keyword embedding distance and sentence embedding distance:
\begin{align}
    f_{\text{sem}}(\mathrm{x}|\mathrm{x}_0) &= f_{\text{sem,key}}(\mathrm{x}|\mathrm{x_0})^P \cdot f_{\text{sem, sen}}(\mathrm{x},\mathrm{x_0})^Q
    \label{eq:sim} \\
    f_{\text{sem,key}}(\mathrm{x}|\mathrm{x_0})&=\min_{e \in \operatorname{keywords}(\mathrm{x_0})} \max_{j} \{\operatorname{cos}(\bm{w}, \bm{e})\}
    \label{eq:keyword-sim} \\
    f_{\text{sem, sen}}(\mathrm{x},\mathrm{x_0}) &= \operatorname{cos}(\bm{x},\bm{x}_0)
    \label{eq:sent-sim}
\end{align}
where $P$ and $Q$ are the relative weight of keyword similarity and sentence similarity, respectively. Here the keywords are extracted by RAKE \cite{rose2010automatic} and the GloVE \cite{pennington2014glove} embeddings are used for computing distance.

The objective function used in UPSA is a multiplicative weighted combination of the aforementioned components. It simultaneously evaluates language fluency, semantic preservation, and lexical diversity:
\begin{equation}
f(\mathrm{x}, \mathrm{x_0}) = f_{\text{flu}}(\mathrm{x})\cdot f_{\text{sem}}(\mathrm{x}|\mathrm{x_0}) \cdot f_{\text{lex}}(\mathrm{x}|\mathrm{x_0})
\label{original_score}
\end{equation}
note that each components are weighted implicitly.
	
To this end, we have reviewed each individual component in the objective function for paraphrase generation. This objective function specifies the requirement for the task of paraphrase generation. Next to discuss is a search algorithm to maximize such objective function approximately by making word-level local edits.

\subsection{Simulated Annealing Search}
The computational cost of exhaustive search renders it infeasible to perform. The number of possible sentences that can be generated combinatorially grows exponentially with the vocabulary size. In other words, an exhaustive search can be done theoretically, but one would need to face the cost of a branching factor of 30,000 (i.e. vocabulary size). Moreover, for the task of paraphrase generation specifically, exhaustive search is not needed as there could be many plausible paraphrases for a sentence. Being able to find an acceptable solution efficiently should suffice.


The general local search algorithm solves discrete optimization problem by starting from a candidate then iteratively moving to neighbours in the solution space by applying local changes, until some criterion is satisfied or the computational budget is exhausted. In other words, every step in the search trajectory itself would be a potential solution. Hence, the main goal is to search efficiently by making moves that follow a good strategy, so that the trajectory can reach at least one high score state. In this dissertation, we use the simulated annealing algorithm described in \cite{liu-etal-2020-unsupervised} as the local search algorithm for solving the optimization problem approximately. Paraphrase generation is a monolingual task that usually have significant overlap between input and output. Hence, the input sentence would serve as a reasonable starting point for the search. 
	
Local search algorithms require a well-defined neighbourhood relation in the solution space for state transition. For the task of paraphrase generation, the solution search space is defined to be all finite length sentences. Search steps are realized by making word-level edit operations. More specifically, each state of the search is represented by: 
\begin{equation}
\mathrm{x_t} = (w_{t,i}|i\in[0,l])
\end{equation}
where $t$ indexes time steps in the search process, and $l$ is the length of the sentence corresponding to state $\mathrm{x_t}$.

Local word-level edits used in this framework are insertion, deletion, and replacement. It can be seen that in theory any arbitrary sentence can be generated from any starting point using these edit operations. Word deletion is straightforward to perform, but word insertion and replacement both require adding appropriate new word to the sentence. The new word added to the sentence are sampled from the same language model used for fluency evaluation. Details will be discussed in rest of this section.

Simulated annealing (SA) is the choice of search algorithm for maximizing the aforementioned objective function \ref{original_score}. The terminology of SA is borrowed from metallurgy, which originally refers to the technique of alternating heating and cooling of a material to produce crystals of desired size while reducing defects. In the context of optimization by stochastic search, this translates to the idea of that the search algorithm would climb to higher function values greedily most of the time, but occasionally allow some non-greedy exploration to escape local optima. SA algorithm explicitly encourages exploration during the initial steps of the search, then becomes more greedy as the search goes on. With this design, the SA algorithm is able to provide reasonable solution even when the search space is large.
\begin{figure}[!t]\centering
    \includegraphics[width=10cm]{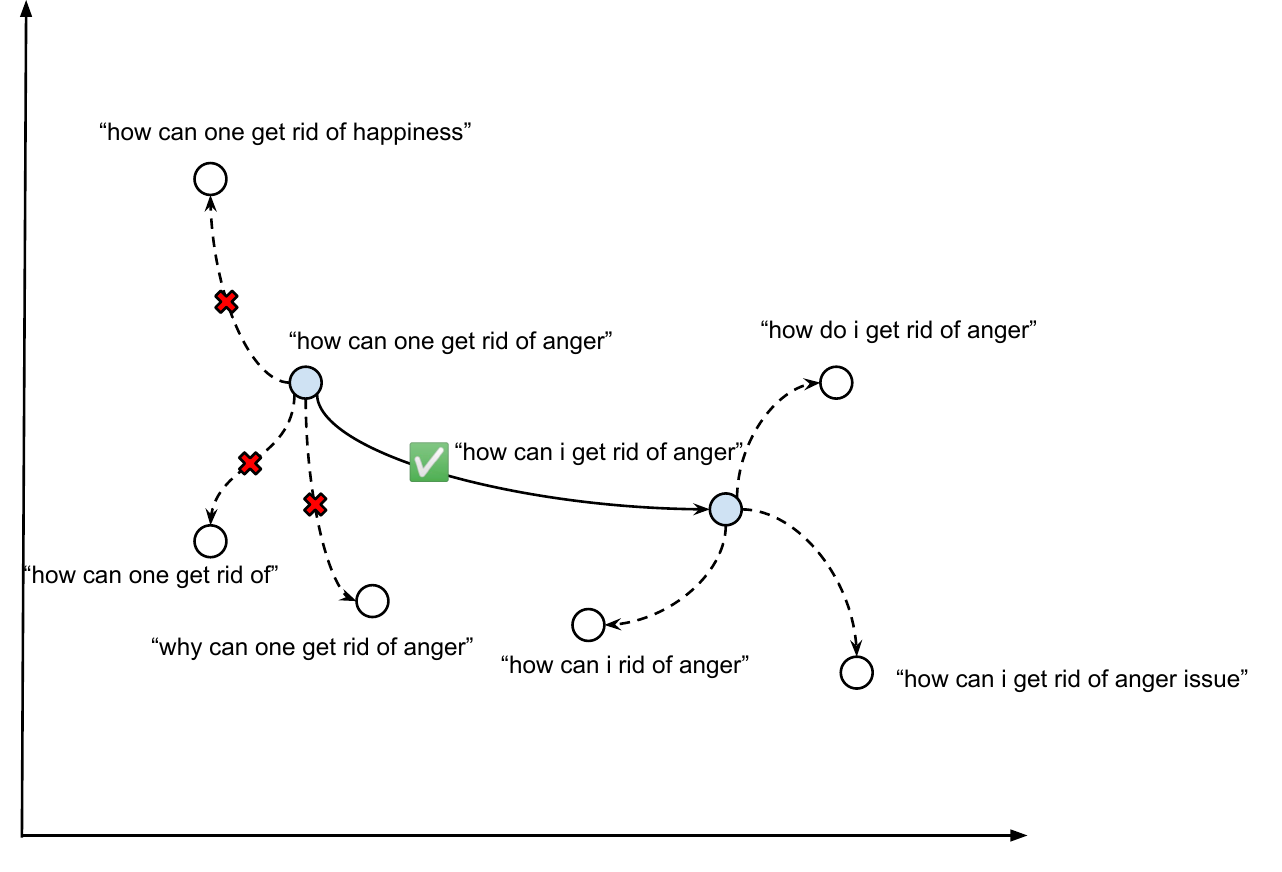}
    \caption{An example of one iteration of simulated annealing search. The red crosses represent rejected proposals, while the green swoosh represents accepted proposal.}
    \label{sentence_space}
\end{figure}

One basic iteration of SA for text generation is done by first making edit proposals, then either accepting or rejecting the proposals. For each iteration, a position to be edited $i$ is sampled from a uniform distribution over the length of the sentence, giving each position an equal chance to be edited. Then a neighbouring state $\mathrm{x_*}$ is proposed based on the ongoing state $\mathrm{x_t}$ at time step $t$ by performing the one of the aforementioned basic edit operations: insertion, replacement, or deletion, at the sampled position of $i$:
\begin{equation}
    i \sim Uniform(0,l)
\end{equation}
where $l$ is the number of words in sentence $\mathrm{x_t}$.

An edit operation from insertion, deletion, and replacement needs to be chosen once the edit position has been determined. Edit operations are also sampled from the three options. More specifically, given probability of $p_{ins}$, $p_{rep}$, $p_{del}$ for insertion, replacement and deletion, respectively, the edit operation is sampled by:
\begin{equation}
 z \sim Categorical(p_{ins}, p_{rep}, p_{del})
\end{equation}
where $z$ is the one-hot vector indicating which edit operation to be performed. In practice, $p_{ins}$, $p_{rep}$, $p_{del}$ are set to equal.

If the sampled edit operation is deletion, then it would be straight forward: given the current state $\mathrm{x_t} = (w_{t,1}, \dots, w_{t,i-1}, w_{t,i}, w_{t,i+1}, \dots, w_{t,l})$ at step $t$ and $i$-th word to be deleted,  the new candidate sentence becomes $\mathrm{x_*} = (w_{t,1}, \dots, w_{t,i-1}, w_{t,i+1}, \dots, w_{t,l})$.

For insertion and replacement, a new word needs to be sampled to add to the sentence. For the appropriate word to be selected, the new candidate word is sampled with probability proportional to the objective score corresponding to the new sentence state induced by the added word
\begin{equation}
    p(w_*|\cdot) =
    \frac{f_{\text{sim}}(\mathrm{x_*}, \mathrm{x_0}) \cdot
    f_{\text{exp}}(\mathrm{x_*}, \mathrm{x_0}) \cdot
    f_{\text{flu}}(\mathrm{x_*})
    }{Z}
    \label{sample_word}
\end{equation}
\begin{equation}
    Z = \sum_{w_* \in \mathcal{W}} f_{\text{sim}}(\mathrm{x_*},   \mathrm{x_0}) \cdot f_{\text{exp}}(\mathrm{x_*}, \mathrm{x_0}) \cdot f_{\text{flu}}(\mathrm{x_*})
\end{equation}
sampling in such way would ensure words that lead to higher scores are more likely to be added.
Due to the expensive computational cost of evaluating Equation \ref{sample_word} for every word in the vocabulary, a forward and backward language model is used to truncate this selection to only top-K words. In other words, only the words that lead to high fluency scores are considered. If the new word is to be placed at the $i$-th slot in sentence $\mathrm{x}_t$, the top-$K$ words being used to evaluate \ref{sample_word} are restricted to:
\begin{equation}
    \mathcal{W}_{t} = \text{top-} K_{w_*} [
    p_{\overrightarrow{\text{LM}}} (w_{t,1}, \dots, w_{t,i-1}, w_*) \cdot
    p_{\overrightarrow{\text{LM}}} (w_*, w_{t,i+1}, \dots, w_{t,l_t})
    ]
    \label{top-k_sampling}
\end{equation}
\sloppy
After the new candidate word is proposed, the new candidate state generated by replacement operation would be $\mathrm{x_*} = (w_{t,1}, \dots, w_{t,i-1}, w_* , w_{t,i+1}, \dots, w_{t,l})$. Likewise, the new candidate state proposed by insertion operation would be $\mathrm{x_*} = (w_{t,1}, \dots, w_{t,i-1}, w_{t,i}, w_* , w_{t,i+1}, \dots, w_{t,l})$.

Copy mechanism is incorporated to preserve named entities and rare key words. These words are usually kept in the sentence during paraphrasing since they carry relatively high amount of important information. However, in practice sentences with these rare words are usually given very low probabilities by language models due to their low uni-gram probabilities.
Hence, the practical issue of these rare words is that they are likely to be deleted or replaced during the search process, but are not likely to be recovered again due to the low language model probabilities of sentences having these words. Inspired by \cite{gu-etal-2016-incorporating}, the copy mechanism is incorporated in the word sampling process by augmenting the top-k candidates with all words in the input sentence. Hence, it would be easier for the model to copy these words from the original sentence. In particular, the truncated candidates become:
\begin{equation}
    \widetilde{\mathcal{W}}_{t} =  \mathcal{W}_{t} \bigcup \{w_{0,1}, \dots, w_{0,l_0}\}
\end{equation}
where $\mathcal{W}_t$ is the top-K words truncated by the forward-backward language model in \ref{top-k_sampling}, and $\{w_{0,1}, \dots, w_{0,l_0}\}$ are the words in the input sentence $\mathrm{x_0}$.

The SA algorithm decides if the new proposal $\mathrm{x_*}$ would be accepted. The essence of SA algorithm is to always accept proposals that lead to higher scores, but lower scored proposals are occasionally accepted as a means to escape local optima. The time-dependent acceptance ratio is given by:
\begin{equation}
    p(\mathrm{accept} | \mathrm{x},\mathrm{x_*},\mathrm{x_0},T) = \operatorname{min}\{1, e^{\frac{f(\mathrm{x_*}|\mathrm{x_0})-f(\mathrm{x}|\mathrm{x_0})}{T}}\}
\label{sa_accept}
\end{equation}
where $T$ is the temperature parameter controlling how likely non-greedy steps are allowed. $T$ is large at the beginning of search then gradually cools down as search goes on. As shown, the probability of accepting a proposal $\mathrm{x_*}$ is always 1 if $f(\mathrm{x_*}|\mathrm{x_0}) > f(\mathrm{x}|\mathrm{x_0})$. Otherwise, the probability of accepting a new proposal $\mathrm{x_*}$ that leads to lower objective score depends on the absolute difference between $f(\mathrm{x_*}|\mathrm{x_0})$ and $f(\mathrm{x}|\mathrm{x_0})$. Earlier steps yield high probability to accept a proposal $\mathrm{x_*}$ when $f(\mathrm{x_*}|\mathrm{x_0}) < f(\mathrm{x}|\mathrm{x_0})$ due to the larger $T$ value. The gradual reduction of temperature in Equation $\ref{sa_accept}$ can be interpreted as the gradual increase in greediness for high objective score. We follow \cite{liu-etal-2020-unsupervised} and set initial temperature to $T_0=3\times10^{-2}$, which is then linearly decreased to zero after a fixed number of search iterations. In this manner, the search algorithm is encouraged to explore a wide region of the search space at the beginning of search. As the search progresses the algorithm commits to an optimum. In theory, with sufficient budget steps for search, the probability of the algorithm settling with a global optimum converges to 1. However, the large search space makes this theoretical guarantee unrealistic. Instead, the main focus of the SA search is to efficiently find a near optimal solution without global inference.
\begin{figure}[!t]\centering
    \includegraphics[width=10cm]{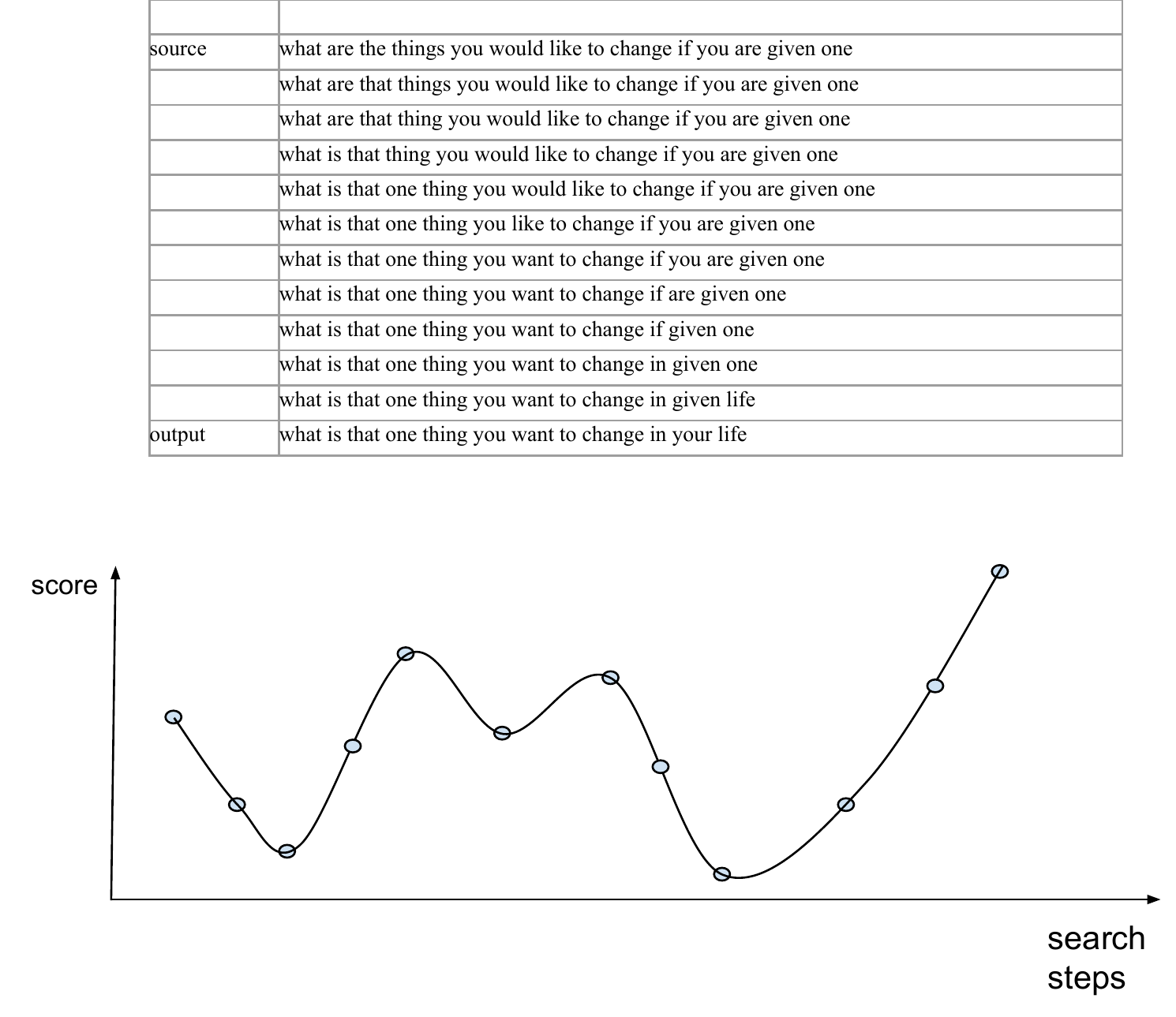}
    \caption{An example search trajectory of paraphrase generation.}
    \label{score}
\end{figure}

\section{Learning From Search}
\label{section: proposed}
One drawback of the search based text generation framework underlies in the manually designed objective function, which is not learnable. Previous work \cite{liu-etal-2020-unsupervised}\cite{Miao_Zhou_Mou_Yan_Li_2019} show that maximizing the objective function by search indeed improves the evaluation metrics (e.g., BLEU and iBLEU) on average for a dataset. Such evidence show that there is indeed a correlation between the manually designed objective function and the true measure of success on a population level. However, we suspect that the heuristic objective defined with a high level of abstraction may not have the granularity to provide accurate guidance when it comes to each individual sentence. Empirical evidence will be shown to support this claim later in experiments section.
\begin{figure}[!t]\centering
    \includegraphics[width=10cm]{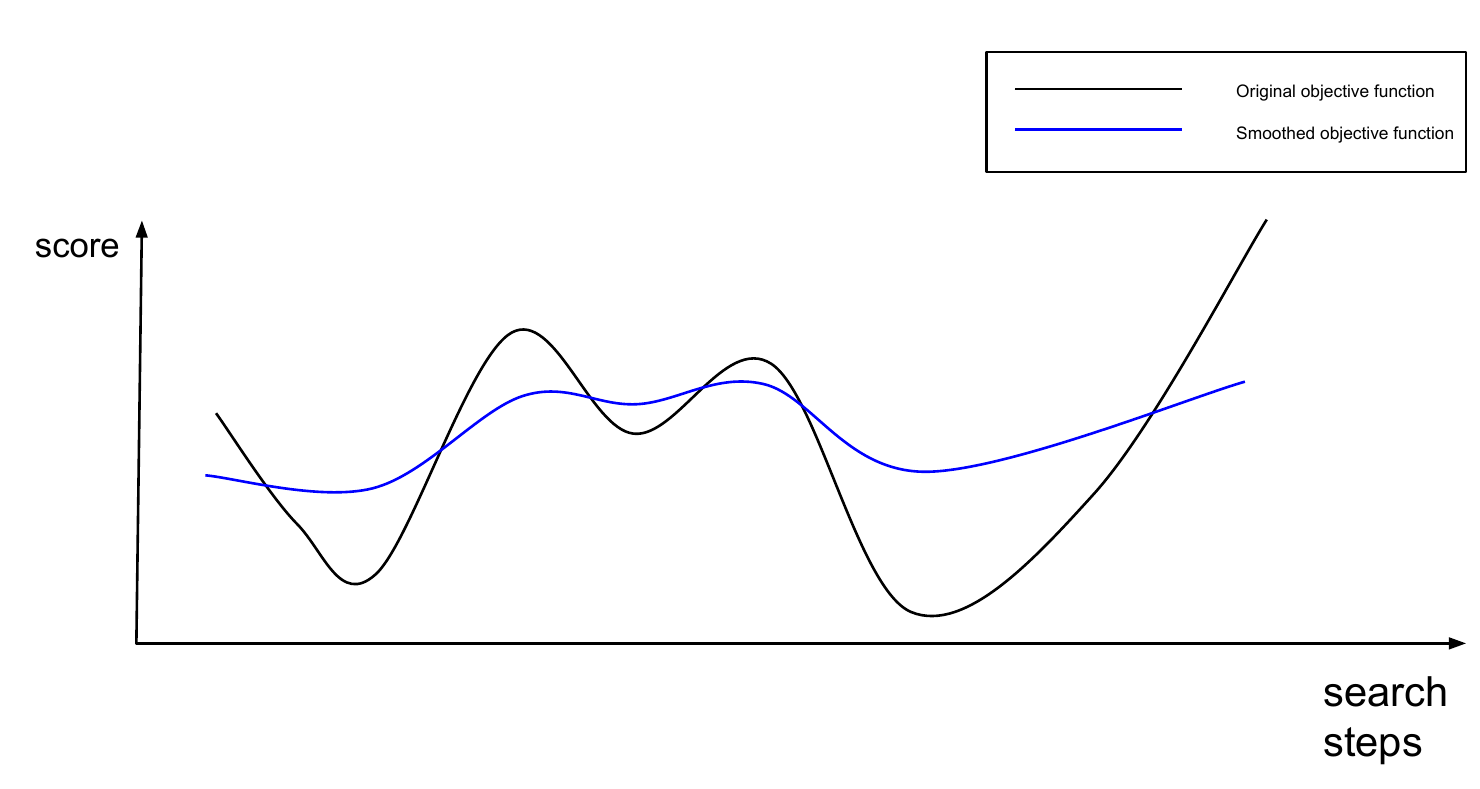}
    \caption{A sketch of our motivation to smooth out the original objective function.}
    \label{sa_smooth}
\end{figure}

Another challenge of local search in general is the trap of local optima. Being trapped in a local optimum that has a relatively higher score than all of its immediate neighbours can leave the search stranded for a very long time. Even though SA algorithm would allow occasionally non-greedy move, the probability of this happening only depends on the absolute difference in the objective scores between two states. In our work, we aim to achieve the goal of learning to make sacrifice by making locally non-greedy actions that are beneficial in the long run. This is accomplished by smoothing the heuristic objective function, so that the jumps in between locally optima are more feasible. As demonstrated in Figure \ref{sa_smooth}, a smoothed objective function would yield smaller difference among the neighbouring region of a state. Such smaller difference would lead to higher acceptance probability for transition to states with lower scores, but with higher scoring regions around them for the following potential steps to reach.

To this end, we propose three approaches for smoothing out the objective function by learning models from the search process of simulated annealing. The learned models are then combined with the original objective function to form a new smoother approximation of the original objective. The learning task is accomplished by two regression models that try to directly predict the scores, and a Seq2seq model that predicts the likelihood of state transitions implicitly. In the following section, we will go through the proposed models, implementation and model tuning details.

\subsection{Value Function}
\label{section: value function}
Our first approach is to directly train a regression model to approximate the objective function, known as value function \cite{rl_value_function}. The motivation of this approach is to directly smooth out the objective function numerically. The procedure of this approach is as follows: 

\begin{enumerate}

\item We perform SA search towards the original objective function, to collect search trajectory samples in the form of $\mathrm{X} = \{\mathrm{x}_i| i \in [0,h]\}$ where $i$ indexes the time steps in a trajectory, and $h$ is the number of search steps in a search trajectory. Each search state in a trajectory is labeled by its own objective score evaluated by the original objective function $f$.

\item The value function network is trained to take as input the original sentence $\mathrm{x}_0$, and a state $\mathrm{x}$ from a search trajectory, then make a prediction of the objective function score $\hat{f}(\mathrm{x}|\mathrm{x}_0)$ for the given state $\mathrm{x}$. The value function network is trained by mean square error (MSE) against the ground-truth objective scores.

\item For inference, the learned value function is directly combined with the original objective $f$, yielding a new objective function $f_{\text{value}}$. This new objective function then guide a next iteration of SA search to generate the final output $\mathrm{x}_T$.
\end{enumerate}

The baseline model used in step (1) to collect search trajectory samples are our own re-implementation of UPSA. We evaluate the performance of this re-implementation to validate the baseline. Detailed results are presented in Chapter 4. We faithfully follow all hyperparameter settings as described in \cite{liu-etal-2020-unsupervised}, setting up a fair comparison between our own models and the original SA search.

In step (2), the value function network $f_v(\mathrm{x}|\mathrm{x}_0)$ is trained to predict the scalar score evaluated by the objective function $f$. As the backbone of this model, a pre-trained \texttt{bert-base-uncased} \cite{devlin-etal-2019-bert} model is adopted and modified for the regression task. The \texttt{bert-base-uncased} model is a transformer \cite{transformer} model pre-trained on Masked Language Modelling (MLM) and Next Sentence Prediction (NSP) with large scale English data. Pre-training enables the model to learn effective latent representation, which can help relevant downstream tasks. This pre-trained model is capable of performing superbly in a variety of tasks \cite{xia-etal-2020-bert} with minimal fine-tuning. Considering this regression task is largely dependent on Natural Language Understanding(NLU), the BERT model is a reasonable choice for our base model. To modify the pre-trained $\texttt{bert-base-uncased}$ model for regression, we extract representation from the pooled output, which is the last layer of the hidden-state of the special leading token of the sequence $[\text{CLS}]$. We then simply add a fully connected layer on top of the pooled output, to generate a scalar prediction of the score. The regression model is then further fine-tuned by mean square error (MSE), in order to minimize the $l2$ error between the score predicted by the value function $f_{v}$ and the ground truth score given by the heuristic objective function $f$.
\begin{equation}
    J_v(f, f_{v})=
\sum_{\mathrm{x} \in \mathrm{X}} (f(\mathrm{x}|\mathrm{x_0})-f_{v}(\mathrm{x}|\mathrm{x_0}))^2
\end{equation}

In the inference stage of step (3), the learned value function prediction $f_v$ is directly combined with the original heuristic objective function $f$ by convex combination, yielding an overall objective function $f_{value}$ as
\begin{equation}
    f_{value}(\cdot | \mathrm{x_0})= k\cdot f_v(\cdot|\mathrm{x_0}) + (1-k) \cdot f(\cdot|\mathrm{x_0})
\end{equation}

\begin{figure}[!t]\centering
    \includegraphics[width=10cm]{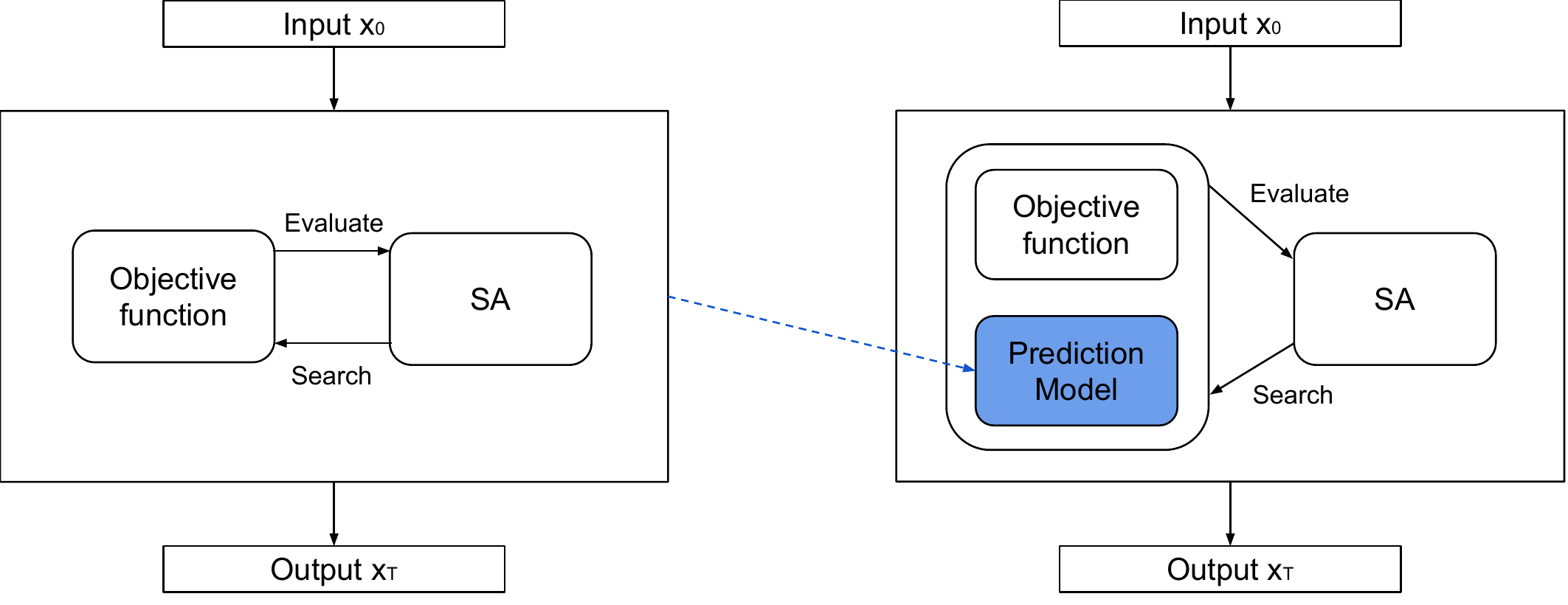}
    \caption{An illustration of our proposed learning from search model. Left hand side is the UPSA framework, which generates trajectory samples for training the prediction model(e.g. value function, Seq2seq model). Then the learned prediction model is combined with the original objective function for heuristic evaluation. Search is performed again with the new combined objective function to generate final output.}
    \label{sa_learn_model}
\end{figure}

\subsection{Max Value Function}
\label{section: max value function}
The second approach we propose is a variant of the value function. Instead of directly learning the objective score of a given state, max value function learns and predicts the maximum objective score of a trajectory starting from any particular state. The idea of this approach is similar to smoothing out the objective function with the value function, but we also want to assign a higher scores to states that lead to subsequent high scoring states in the search trajectory. In other words, the goal of the max value function is to learn to sacrifice immediate scores in order to reach even higher scores later in the search trajectory. Max value function is trained and used for inference almost the same way as the value function, except that in step (1), each state is labeled by the maximum objective score that follows that particular state in the search trajectory. Training objective in step (2) is to learn and predict these new pseudo-labels. Inference in step (3) is the same as the value function approach, using a convex combination of the original score $f$ and the new score $f_{v*}$. We use the exact same $\texttt{bert-base-uncased}$ pre-trained model with an added fully connected regression layer on top of the pooled output. Loss function for training the max value function is also mean square error(MSE):
\begin{equation}
    J_{v*}(f, f_{v*})= \sum_{\mathrm{x} \in \mathrm{X}} [f_{v*}(\mathrm{x}|\mathrm{x_0})-\max_{\mathrm{x'}\in \mathrm{X}}f(\mathrm{x'}|\mathrm{x_0}) ]^2
\end{equation}
here $\mathrm{x}'$ is the highest scoring state that follows the particular state $\mathrm{x}$ in trajectory $\mathrm{X}$.

\subsection{Seq2seq Probabilistic Model}
\label{section: seq2seq}
The third approach uses a Seq2seq model to smooth out the objective function. This idea is similar to \cite{NEURIPS2020_7a677bb4}, in which they also smooth out the noise in the searching and learning frame work using a Seq2seq model. Unlike the regression models used in the value function and max value function, this approach uses the emission probability of a Seq2seq model for smoothing effect. More specifically, we train a seq2seq model which takes as input the original input, and predicts the search output (i.e. paraphrases). During inference, the predicted probability of a particular state to be emitted from the seq2seq model given the original input sentence is combined with the original objective function to provide a smoothing effect. The procedure of the seq2seq smoothing approach is as follows

\begin{enumerate}
    \item We perform SA search towards the original objective function, to collect search trajectory samples $\mathrm{X} = \{\mathrm{x}_i| i \in [0,h]\}$ similar to the value function approaches. Then each trajectory would generate one pseudo-parallel label in the form of $(\mathrm{x_0}, \mathrm{x_T})$. Each pseudo-parallel training example $(\mathrm{x_0}, \mathrm{x_T})$ consists of the original input sentence $\mathrm{x_0}$ and the outcome of search $\mathrm{x_T}$ (i.e. the final state in search trajectory).
    
    \item The seq2seq model is trained to take the input sentence $\mathrm{x_0}$ of a trajectory and predict the search output $\mathrm{x_T}$. The training objective is to minimize the cross-entropy loss against the pseudo-labels from step (1).
    
    \item For inference, the probability of emitting a particular sentence $P_{i,v}^{\text{(s2s)}}$ is directly combined with the original objective $f$, yielding a new objective function $f_\text{seq2seq}$. This new objective function is used by the SA algorithm to search again and output the final sentence $\mathrm{x}$.
    
\end{enumerate}
To train a Seq2seq model using the pseudo-label from search, we train a state-of-the-art transformer-based Seq2seq model $P_{i,v}^{\text{(s2s)}}$ by cross-entropy loss:
\begin{equation}
    J_{\text{CE}} = - 
    \sum\nolimits_{i=1}^{l}
    \sum\nolimits_{v \in \mathcal{V}}
    w_{i,v} \ \mathrm{log} \ P_{i,v}^{\text{(s2s)}}
\end{equation}
Where $w_{i,v}$ is the binary value indicating whether the $i$-th word is $v$ or not in the search output, and $P_{i,v}^{\text{(s2s)}}=P_{\text{s2s}}(w_i=v|w_{<i}, \mathrm{x_0})$ is predicted by the Seq2seq model.

The learned Seq2seq model $P_\text{(s2s)}$ is combined with the original heuristic objective function $f$ by:
\begin{equation}
    f_{\text{s2s}}(\cdot|\mathrm{x_0}) = k\cdot d\cdot P_{\text{s2s}}(\cdot|\mathrm{x_0}) + (1-k) \cdot f(\cdot|\mathrm{x_0})
\end{equation}
\begin{equation}
    P_{\text{s2s}}(\mathrm{x}|\mathrm{x_0}) = \prod\nolimits_{i=1}^{L} p(w_i|w_{<i}, \mathrm{x_0})
\end{equation}
Here, d is a scaling factor (set to 100) to scale a probabilities to a similar range of Equation \ref{original_score}. $k$ is a relative weighting hyperparameter for the two terms. $f_{\text{s2s}}$ becomes the new objective for generating final outputs by another iteration of SA search.

Although both using a Seq2seq model to learn from the pseudo-parallel search output, our work differs from \cite{NEURIPS2020_7a677bb4}, in which  $P_{\text{s2s}}$ is used to directly generate output tokens in an auto-regressive manner. Instead, our work aims to improve the quality of generated text by improving the original heuristic objective function. Our work provides insight on the search objective, which is an important building block for all search-based text generation frameworks. Moreover, we want to answer the curious question of whether search guided by a heuristic objective function can help improve the objective function in a bootstrapping manner without supervision.




%% file: tex/experiments.tex
\chapter{Experiments}

\section{Dataset}
We follow \cite{liu-etal-2020-unsupervised} and evaluate our models on unsupervised paraphrase generation with Quora question pair dataset\footnote{Quora question pair dataset\url{https://www.kaggle.com/c/quora-question-pairs}}, which is a collection of question pairs that are identified as duplicate questions. These duplicated questions can be interpreted as paraphrases for each others. Each data instance consists of its own id, two questions with their respective unique id numbers, as well as a target variable indicating whether they are duplicate questions.
Preprocessing of the data involves lower-casing, removing punctuation, and finally deduplicating to avoid any particular instance being assigned more weight than others in training and evaluation. After cleaning up the data, we reserve 10K and 20K of parallel paraphrase pairs for validation and test, respectively. The remaining 500K non-parallel sentences are used for training language model (for fluency score and proposal truncation) as well as collecting search trajectories for training the value function, max value function, Seq2seq model.

\section{Settings}
\subsection{Search Algorithm Parameters}
We faithfully follow the hyperparameter settings of the simulated annealing search algorithm as in UPSA \cite{liu-etal-2020-unsupervised} for two reasons: hyperparameters in UPSA are tuned by grid search and validated on the same Quora dataset, which already achieved state-the-art performance in paraphrase generation; we will use this set of hyperparameter settings as the basis of our ablation study, which sets up a fair comparison with the original UPSA model. In particular, initial temperature $T_\text{init}$ is set to $3 \times 10^{-2}$, then decay according to $T = \operatorname{max}\{T_\text{init} - C \cdot t, 0\}$, where the annealing rate $C$ is set to $3 \times 10^{-4}$. In this way temperature $T$ will drop to zero linearly in 100 iterations. The language model for word proposal and evaluating fluency score $f_{flu}$ is parameterized by a two-layer LSTM model with 300 hidden units. Training of the language model uses only the non-parallel training split of the Quora dataset. The relative weight of each component in the heuristic objective $P, Q$ , and $S$ are set to $8,1,1$, respectively.

\subsection{Search Trajectories Collection}
To learn from the SA search dynamics, we perform SA search on the 50K inputs from the training set, generating 50K search trajectories totalling 1.3M search steps for training our proposed models. Each state visited in search trajectories becomes a training data point, which consists of its own trajectory id, search step index, edit operation used, and score evaluated by the original objective. To train the value function $f_v$, each data instance is cleaned up to keep the original input sentence $\mathrm{x_0}$, the sentence corresponding to the state at the particular search step $\mathrm{x_t}$, and the objective score $f(\mathrm{x_t})$. For training of max value function $f_{v*}$, the objective score is set to the highest objective score in the trajectory after each particular search step. To generate the pseudo-parallel labels for training Seq2seq model $P_{\text{s2s}}$, each search trajectory would contribute exactly one training sample, consisting of only the original input sentence $\mathrm{x_0}$ and the final output sentence $\mathrm{x_T}$ from search.

\subsection{Model Architecture and Tuning}
Both of value function $f_v$ and max value function $f_{v*}$ adopt the pre-trained \texttt{bert-base-uncased} model by directly adding an extra fully-connected regression layer to the pooled output of the \texttt{[CLS]} token, outputting a single scalar prediction. The two models are both further fine-tuned by 50 epochs, with batch size of 64. After tuning with the validation split of the Quora dataset, we choose a scheduler that linearly warms the learning rate up to $1.5e-6$ from 0 in 10 epochs, followed by a cosine decay period.

Our Transformer-based Seq2seq model has 3 layers, 8 heads, and 512 hidden units. The model is trained from scratch by 10 epochs with a batch size of 64, using the same cosine learning rate scheduler as the value function and max value function.

\section{Evaluation Metrics}
To set up a fair comparison with other state-of-the-art methods for unsupervised paraphrase generation, we adopt the standard BLEU and iBLEU as evaluation metrics. BLEU \cite{papineni-etal-2002-bleu} measures length-penalized $n$-gram overlaps between the output and the ground-truth. iBLEU is a variant of BLEU that explicitly discounts $n$-gram overlap between the input and output, to favor lexical dissimilarity between input and output in the paraphrase generation task \cite{sun-zhou-2012-joint}. Both BLEU and iBLEU take value between 0 and 1. A larger BLEU or iBLEU score indicates higher $n$-gram similarity with the reference. For iBLEU specifically, a higher score also corresponds to higher dissimilarity with the input sentence. In the case of a BLEU score of 1, that would mean the generated output is identical to the reference. If the iBLEU is evaluated to be 1, that means the generated output is identical to the reference, while completely different from the input.

\begin{table}
\centering
\footnotesize
\begin{tabular}{lrl}
\hline \textbf{Reported in \cite{liu-etal-2020-unsupervised}} & \textbf{BLEU} & \textbf{iBLEU} \\ \hline
VAE & 13.96 & 8.16 \\
CGMH & 15.73 & 9.94 \\
SA & 18.21 & 12.03 \\
\hline \textbf{Our implementations}\\ \hline
UPSA & 18.16 & 12.40 \\
UPSA+Value & 18.67 & 13.11 \\
UPSA+MaxValue & 18.81 & 13.12 \\
UPSA+Seq2SeqProb & \textbf{20.20} & \textbf{13.42} \\
\hline
\end{tabular}
\caption{\label{performance} Unsupervised paraphrasing results. BLEU and iBLEU scores are presented in percentage format.}
\end{table}

\section{Results of Unsupervised Paraphrase Generation}
Table \ref{performance} presents the result of automatic evaluation for unsupervised paraphrase generation. Experimental results show that our implementation of UPSA reproduces similar performance as reported in \cite{liu-etal-2020-unsupervised}, verifying that our implementation is correct and setting up the foundation for our study. Moreover, experimental results show that unsupervised paraphrasing by simulated annealing outperforms variational sampling (VAE, \cite{bowman-etal-2016-generating}) and Metropolis-Hasting sampling (CGMH, \cite{Miao_Zhou_Mou_Yan_Li_2019}) in both BLEU and iBLEU, showing that UPSA is a competitive model for unsupervised paraphrase generation.

It can be seen from Table \ref{performance} that UPSA boosted by any of the three of our methods consistently outperforms the original UPSA in terms of both BLEU and iBLEU. This shows that our attempt to smoothen the heuristic objective function indeed improved the search performance.

In particular, using the Seq2seq model’s probability achieves the best performance among all three of our approaches, yielding an improvement of 2.04 BLEU and 1.02 iBLEU over the original SA search. We attribute this improvement to the learned probability being heterogeneous from the original heuristic objective function, thus providing the strongest smoothing effect.

\begin{figure}[!t]\centering
    \includegraphics[width=10cm]{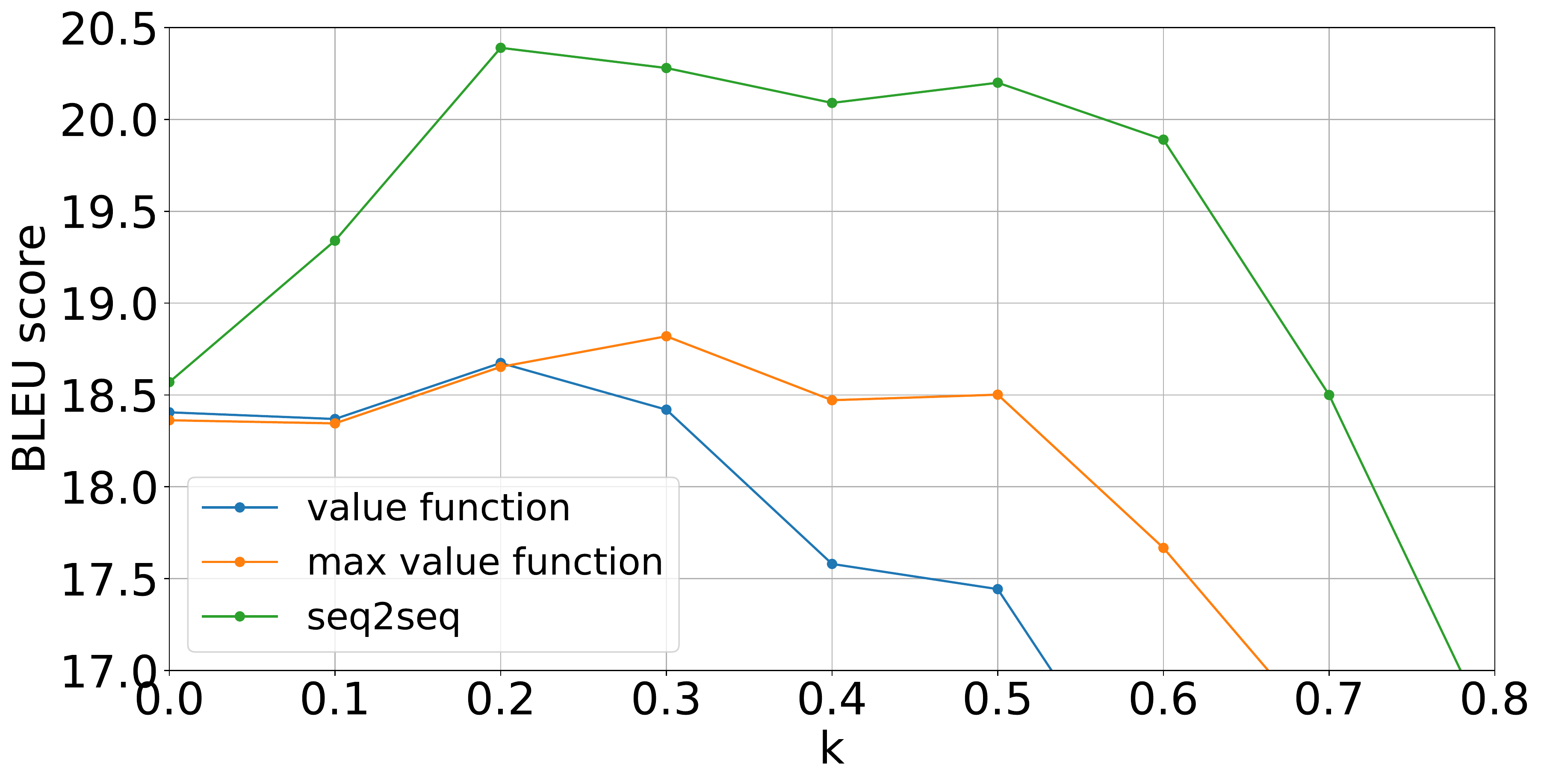}
    \includegraphics[width=10cm]{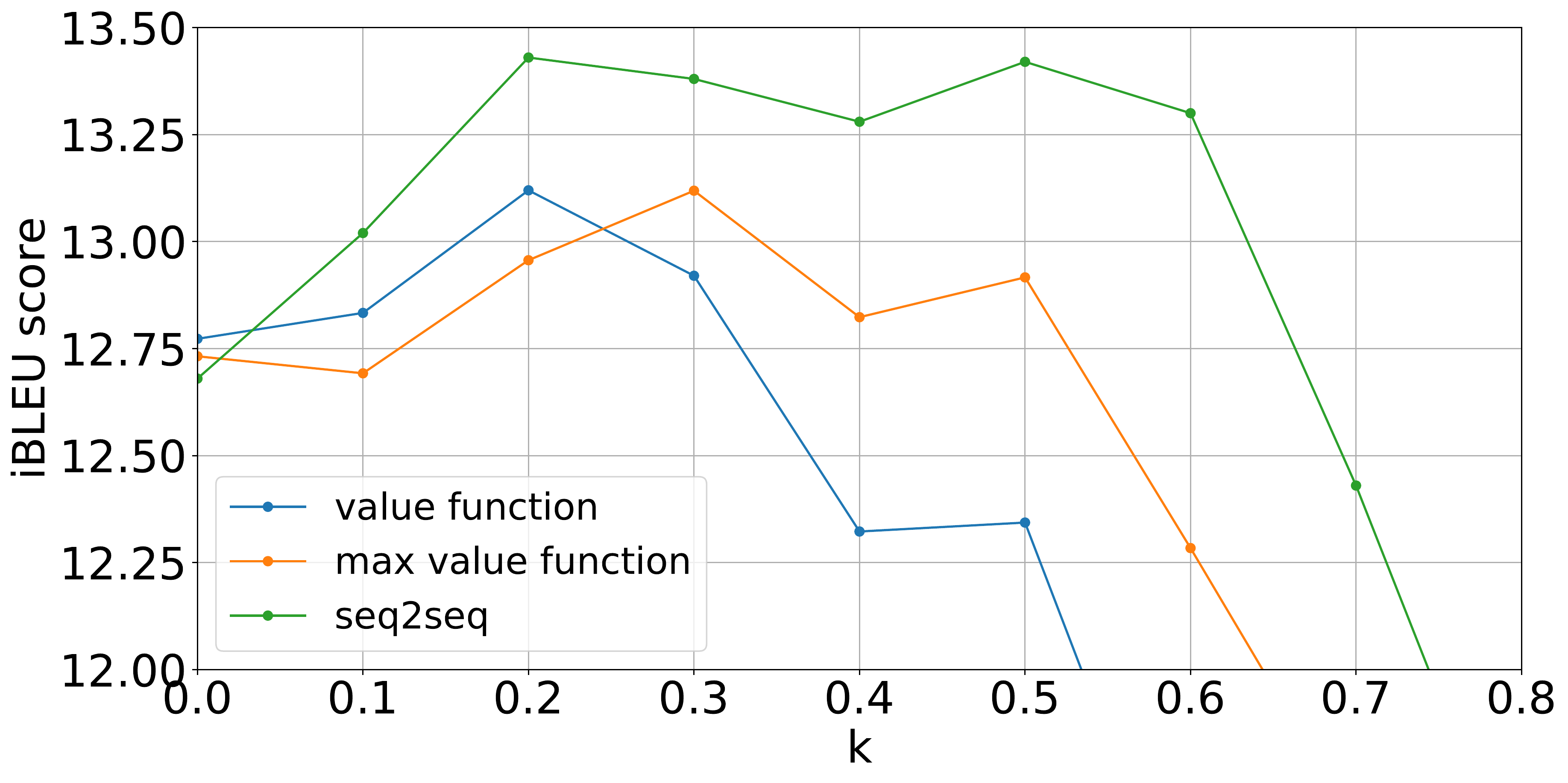}
    \caption{BLEU (upper) and iBLEU (lower) versus the relative weight $k$.}
\label{weight_plot}
\end{figure}

We further show how the performance varies with the relative weight of the learned models in the search objective in Figure \ref{weight_plot}. As seen, all of our models exhibit a similar trend: the performance increases when the learned models are combined with the objective function $f$ by a small weight. This shows that all of our proposed models are able to learn and model search dynamic to a reasonable extent, and thus improving the search. More specifically, max value function $f_{v*}$ outperforms value function $f_v$ when guiding the search, showing our intuition of learning to sacrifice is indeed effective. However, performance of all three of our models decrease when the learned models are weighted more than the original objective function. Furthermore, if the original objective is ignored and search is guided only by any of the learn model only, performance drastically decreases. This shows that our learned models are skewed from the original objective function, rendering them inadequate for guiding the search on their own. Note that when combining the learned models with original objective function, weight $k$ for all models and the scaling factor $d$ for Seq2seq are the only tunable hyperparameters. After training our models, they can immediately boost search performance without excessive tuning. This shows our idea of learning to model the search dynamic is effective. 

\section{Analysis}
Besides the standard BLEU and iBLEU evaluation metrics, we also include several qualitative and quantitative analyses to demonstrate the insight on what our models learn and why they can improve the search-based text generation framework. These analyses are done by observing the difference in search dynamics with and without the proposed models.

\subsection{Correlation of Objective Function with Evaluation Metrics}
One of the key assumptions for the search-based framework to be feasible is that the manually designed objective function correlates with the true measures of success, BLEU and iBLEU, on a population level. To investigate whether this is indeed true, we measure the point-wise correlation between the heuristic objective and BLEU or iBLEU. Note that our models aim to smooth out the objective function, which intuitively should improve the correlation due to the removal of noise. Hence, we also measure the correlation between the measures of success (i.e., BLEU and iBLEU) and the new objective function given by our value function, max value function, and Seq2seq model.

To get an empirical estimate of the correlation, we randomly sampled 1000 sentence input to perform SA search with original objective function. Each input would generate a search trajectory with multiple states visited. Then each search step in trajectories is re-evaluated by the new objective function given by our learned models. We adopt the Spearman correlation coefficient $\rho$, which is a rank correlation metric assessing how well a monotonic function can describe the relationship between two variables. This measure of correlation is appropriate for our study since the acceptance probability of proposal depends only on the absolute difference between old and new objective scores. The Spearman correlation between objective scores and BLEU $\rho_{\text{obj}, \text{BLEU}}$ is computed as follows:
\begin{enumerate}
    \item Given a set of search trajectories, we re-evaluate the objective scores and BLEU/iBLEU for each state visited. We then compute the ranking variable $\text{rg}_\text{obj}$ for each objective score, and the ranking variable $\text{rg}_\text{BLEU}$ for each BLEU score.
    
    \item Compute the Pearson correlation coefficient between $\text{rg}_\text{obj}$ and $\text{rg}_\text{BLEU}$ as follows
    
    \begin{equation}
         \rho_{\text{obj}, \text{BLEU}}
        = \frac{\operatorname{cov}(\text{rg}_\text{obj},  \text{rg}_\text{BLEU})}{\sigma_{\text{rg}_\text{obj}} \sigma_{\text{rg}_\text{BLEU}}}
    \end{equation}
    where $\operatorname{cov}(\text{rg}_\text{obj}, \text{rg}_\text{BLEU})$ is the covariance matrix of the rank variables, and $\sigma_{\text{rg}_\text{obj}}$, $\sigma_{\text{rg}_\text{BLEU}}$ are standard deviations of the rank variables, respectively.
    
    \item $\rho_{\text{obj}, \text{iBLEU}}$ can be computed similarly. 
\end{enumerate}

As seen, the Spearman correlation takes a real value in the range of $[0,1]$. A higher value of $\rho$ indicates more similar ranking between the two variables.

\begin{table}
\centering
\footnotesize
\begin{tabular}{ |p{3cm}||p{3cm}||p{3cm}||p{3cm}|  }
    \hline
    \textbf{} & \multicolumn{3}{|c|}{\textbf{Correlation between objective scores and BLEU}} \\
    \hline
    \textbf{Learned model weight} & \textbf{Seq2seq} & \textbf{Value function} & \textbf{Max value function} \\
    \hline
    \textbf{k=0 (original score)} & 0.272229 & 0.272229 & 0.272229 \\
    \hline
    \textbf{k=0.1} & 0.369467 & 0.266385 & 0.272365 \\
    \hline
    \textbf{k=0.2} & 0.430051 & 0.259276 & 0.271866 \\
    \hline
    \textbf{k=0.3} & 0.498687 & 0.250833 & 0.270353 \\
    \hline
    \textbf{k=0.4} & 0.547690 & 0.240596 & 0.267475 \\
    \hline
    \textbf{k=0.5} & 0.586287 & 0.227615 & 0.262661 \\
    \hline
    \textbf{k=0.6} & 0.616134 & 0.210865 & 0.254570 \\
    \hline
    \textbf{k=0.7} & 0.647185 & 0.188805 & 0.241569 \\
    \hline
    \textbf{k=0.8} & 0.665456 & 0.159719 & 0.222038 \\
    \hline
    \textbf{k=0.9} & 0.696291 & 0.122843 & 0.193834 \\
    \hline
    \textbf{k=1 (learned model only)} & 0.713200 & 0.077553 & 0.155707 \\
    \hline
\end{tabular}
\caption{\label{correlation BLEU} Correlation between objective scores and BLEU.}
\end{table}

\begin{table}
\centering
\footnotesize
\begin{tabular}{ |p{3cm}||p{3cm}||p{3cm}||p{3cm}|  }
    \hline
    \textbf{} & \multicolumn{3}{|c|}{\textbf{Correlation between objective scores and iBLEU}} \\
    \hline
    \textbf{Learned model weight} & \textbf{Seq2seq} & \textbf{Value function} & \textbf{Max value function} \\
    \hline
    \textbf{k=0 (original score)} & 0.271930 & 0.271930 & 0.271930 \\
    \hline
    \textbf{k=0.1} & 0.359856 & 0.266143 & 0.272092 \\
    \hline
    \textbf{k=0.2} & 0.420080 & 0.259099 & 0.271619 \\
    \hline
    \textbf{k=0.3} & 0.488268 & 0.250733 & 0.270139 \\
    \hline
    \textbf{k=0.4} & 0.536743 & 0.240584 & 0.267302 \\
    \hline
    \textbf{k=0.5} & 0.574768 & 0.227697 & 0.262534 \\
    \hline
    \textbf{k=0.6} & 0.604170 & 0.211048 & 0.254493 \\
    \hline
    \textbf{k=0.7} & 0.635022 & 0.189097 & 0.241544 \\
    \hline
    \textbf{k=0.8} & 0.653477 & 0.160125 & 0.222065 \\
    \hline
    \textbf{k=0.9} & 0.685465 & 0.123363 & 0.193915 \\
    \hline
    \textbf{k=1 (learned model only)} & 0.704542 & 0.078181 & 0.155833 \\
    \hline
\end{tabular}
\caption{\label{correlation iBLEU} Correlation between objective scores and iBLEU.}
\end{table}

Table \ref{correlation BLEU} and \ref{correlation iBLEU} show the Spearman correlation coefficient of objective function combined with three of our learned models, respectively. It can be seen that in fact the rank correlation between the original objective function and true measures of success is rather low. Moreover, due to the similar formula of BLEU and iBLEU, the two tables present similar patterns: the learned Seq2seq model combined with the original objective function increases correlation with BLEU and iBLEU monotonically as the weight $k$ increases. However, this monotonic increase in correlation with BLEU and iBLEU does not exactly correspond to increase in performance when the new objective function is used. Instead, a small weight given to the learned Seq2seq probability yields the best performance. This shows our Seq2seq model is able to smooth out some noise in the original objective. However, the learned model is not exactly aligned with the original objective function, leading to the completely degenerated performance when it guides the search by itself. On the other hand, the objective function combined with value function and max value function consistently decrease the correlation between objective score and BLEU or iBLEU as the weight $k$ increases. The decline of such correlation does not show a consistent relationship with changes in performance measured by BLEU and iBLEU. This shows that our value function and max value function approaches do not necessarily smooth out the objective function, even though they are still able to improve search performance. However, due to the lack of smoothing effect, such improvement is lower compared to that from the Seq2seq model.

\begin{figure}
  \centering

  \subfloat[BLEU with Seq2seq model ]{\label{objective_metrics:1}\includegraphics[width=80mm]{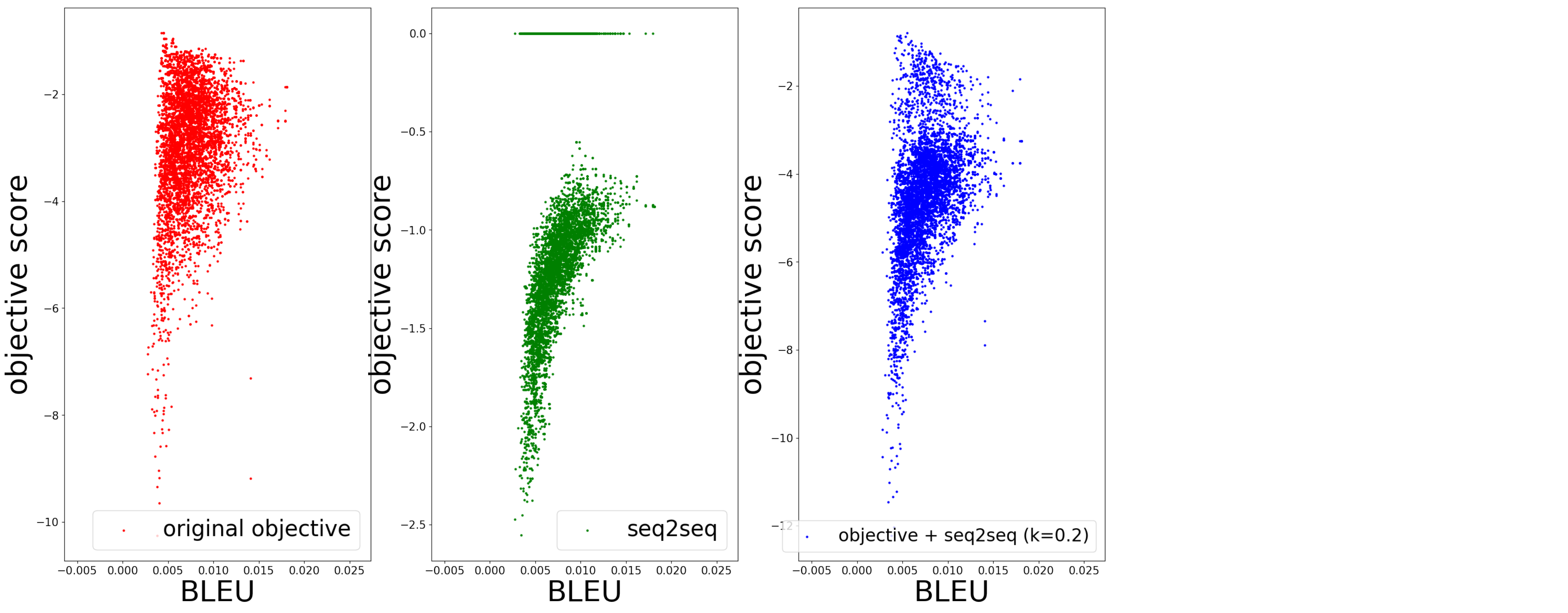}}
  \subfloat[iBLEU with Seq2seq model]{\label{objective_metrics:2}\includegraphics[width=80mm]{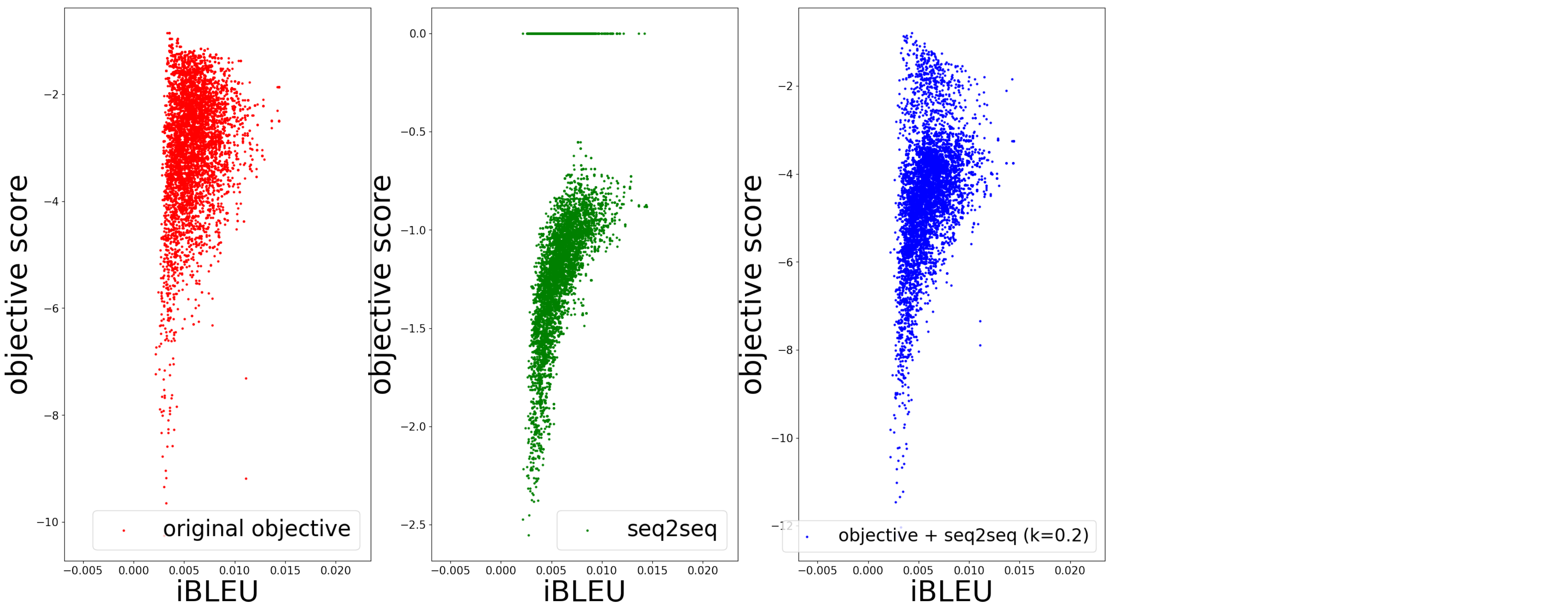}}
  \\
  \subfloat[BLEU with value function]{\label{objective_metrics:3}\includegraphics[width=80mm]{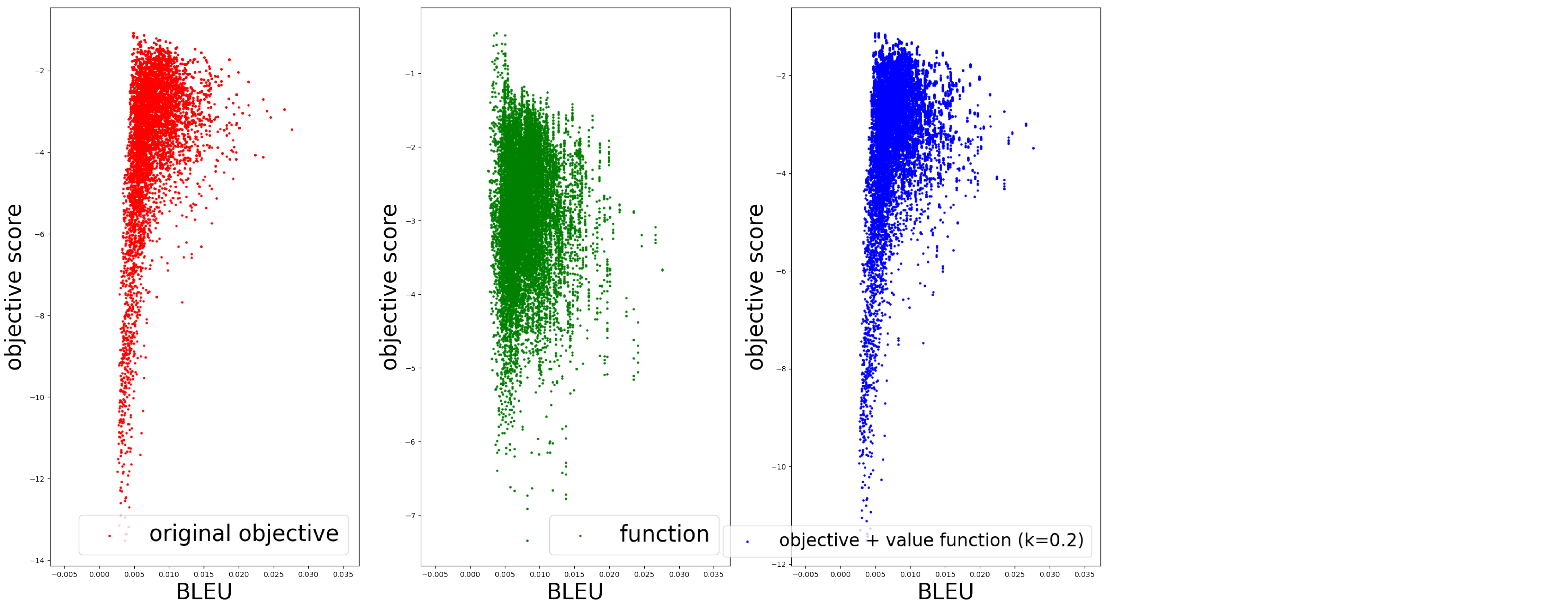}}
  \subfloat[iBLEU with value function]{\label{objective_metrics:4}\includegraphics[width=80mm]{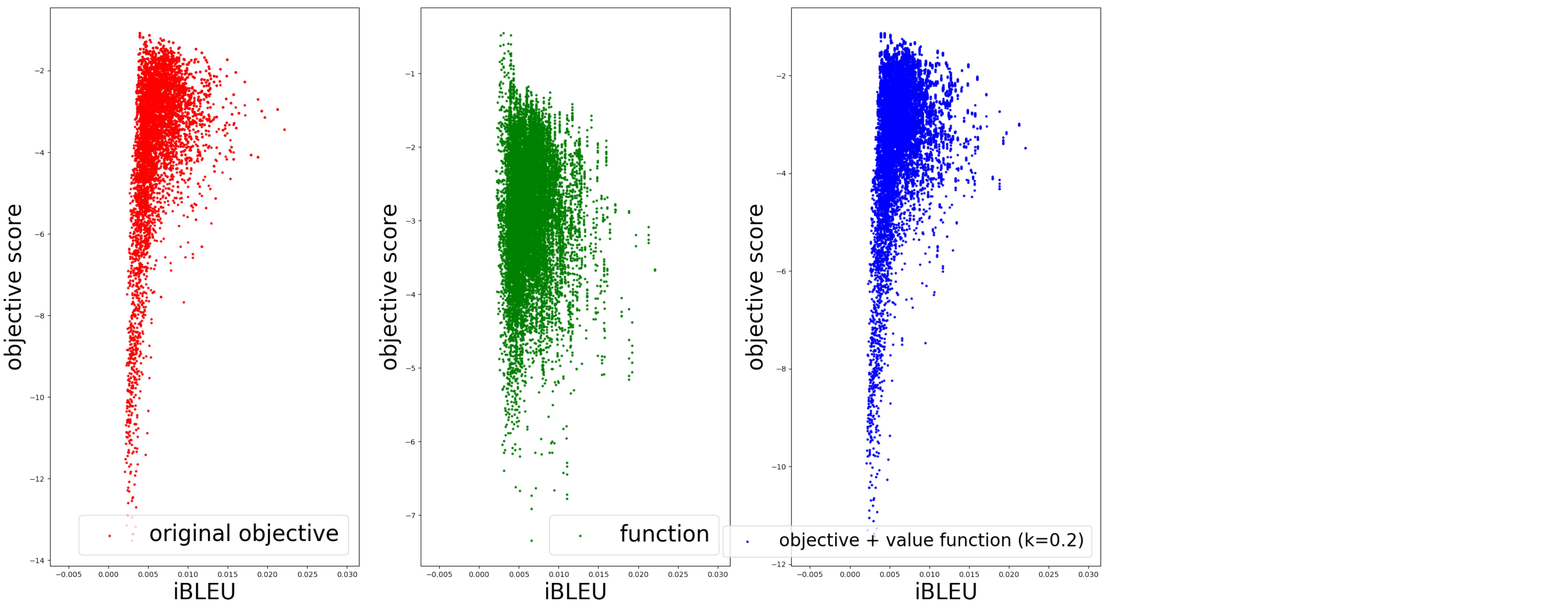}}
  \\
  \subfloat[BLEU with max value function]{\label{objective_metrics:5}\includegraphics[width=80mm]{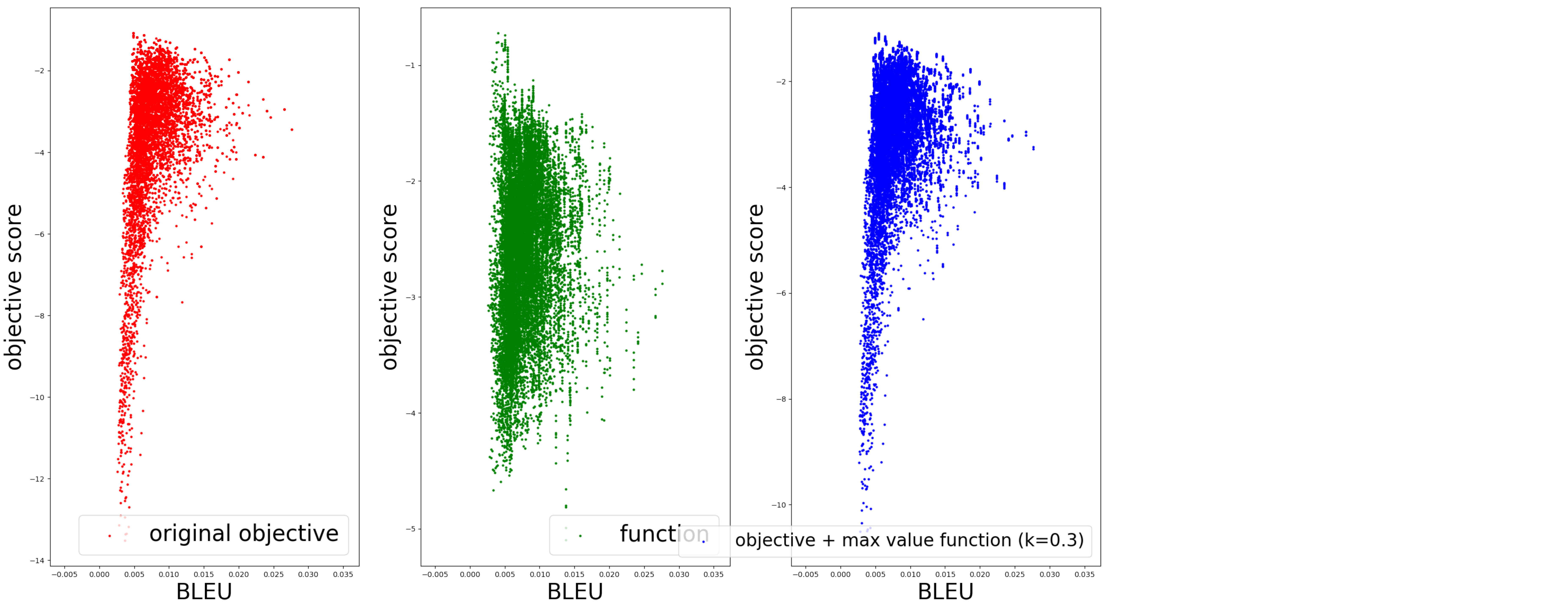}}
  \subfloat[iBLEU with max value function]{\label{objective_metrics:6}\includegraphics[width=80mm]{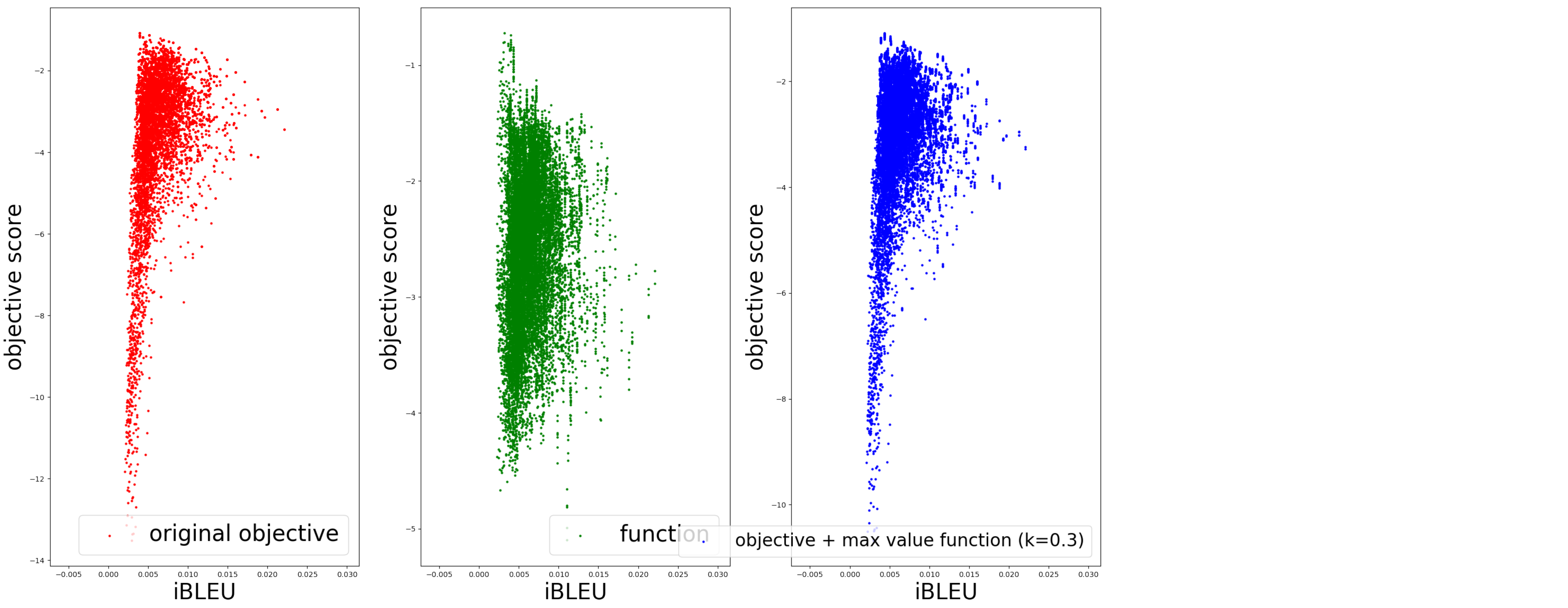}}
  
  \caption{visualization of the mapping between objective scores (vertical axes) and measure of success (BLEU and iBLEU, horizontal axes). Red plots are the original score; green plots are score predicted by the value function, max value function, or Seq2seq model; Blue dots are new objective score combined with corresponding model. }
  \label{objective_metrics}
\end{figure}

Figure \ref{objective_metrics} presents visualization of the mapping between objective scores and measure of success (BLEU and iBLEU). We can see that the Seq2seq model smooth out the objective function by skewing the objective scores. However, the value function and max value function does not seem to have such an effect. Specifically, from Figure \ref{objective_metrics:1} and Figure \ref{objective_metrics:2} we observe that the Seq2seq model combined with the original objective function would pull the objective score down at the lower BLEU or iBLEU region, meaning that the learned model effectively lowers the objective score of some falsely high scored instances when the measure of success (BLEU and iBLEU) is in fact very low. It can also be seen from Figure \ref{objective_metrics:1} and Figure \ref{objective_metrics:2} that the learned Seq2seq model seems to skew the objective score towards a monotonically increasing linear function between the objective score and measure of success (BLEU and iBLEU), which is consistent with our previous conclusion that the learned Seq2seq model increases the linear correlation of objective score with BLEU and iBLEU. The value function and max value function, on the other hand, both learn a more flat-tailed Gaussian-like mapping between the objective score and the BLEU or iBLEU.


\subsection{Acceptance Ratio}
All three of our models play the role of “surrogate” objective function, which governs the search by changing the acceptance probability of given proposals. Hence, we investigate how the acceptance ratio changes with the new objective function modified by our proposed models. Furthermore, we are interested in whether there is a quantitative relationship between the acceptance ratio and the measures of success (i.e., BLEU and iBLEU).

To measure the acceptance ratio, we randomly sampled 1000 input sentence from the test set, then perform search with the objective function combined with Seq2seq, value function, and max value function with various weight. An empirical estimate of proposal acceptance ratio is done by counting the average number of accepted proposals out of 100 sampling steps for each model, which is equivalent to the trajectory length given fixed sampling steps. Table \ref{trajectory length} presents the trajectory length per 100 sampling steps for objective function combined with Seq2seq model, value function, and max value function, respectively with varying weight.

As seen, when searching towards the adjusted objective functions, increasing the weight of Seq2seq model would increase the average acceptance ratio consistently, indicating more edit operations are being realized leading to more diverse output. Different from the Seq2seq model, both the value function and max value function show patterns of decreasing acceptance ratio only when given a small weight ($k < 0.7$), but acceptance ratio increases when weight is large ($k > 0.8$).
Note that for all three models the optimal weight $k$ for the best BLEU and iBLEU is in the range of $[0.2,0.3]$. With a weight in this range Seq2seq roughly preserve the same acceptance ratio of the original objective function, while both value function and max value function decreases the acceptance ratio. This shows that the Seq2seq model when combined with the original objective function does not blindly increase of decrease the acceptance probability on average, but rather makes more informed selection when evaluating acceptance probability. The two regression models, on the other hand, are keen on keeping the original content in the sentence.


\begin{table}
\centering
\footnotesize
\begin{tabular}{ |p{3cm}||p{3cm}||p{3cm}||p{3cm}|  }
    \hline
    \textbf{} & \multicolumn{3}{|c|}{\textbf{Trajectory length (per 100 sampling steps)}} \\
    \hline
    \textbf{Learned model weight} & \textbf{Seq2seq} & \textbf{Value function} & \textbf{Max value function} \\
    \hline
    \textbf{k=0 (original score)} & 21.1071 & 21.1071 & 21.1071 \\
    \hline
    \textbf{k=0.1} & 21.2125 & 18.1641 & 18.1611 \\
    \hline
    \textbf{k=0.2} & 21.2688 & 17.4694 & 17.3814 \\
    \hline
    \textbf{k=0.3} & 21.3222 & 17.0991 & 17.2432 \\
    \hline
    \textbf{k=0.4} & 21.4179 & 17.4084 & 16.7757 \\
    \hline
    \textbf{k=0.5} & 21.6449 & 17.4724 & 16.3543 \\
    \hline
    \textbf{k=0.6} & 21.9042 & 17.7417 & 17.3984 \\
    \hline
    \textbf{k=0.7} & 22.2926 & 18.5505 & 18.1631 \\
    \hline
    \textbf{k=0.8} & 23.2326 & 20.006 & 20.3043 \\
    \hline
    \textbf{k=0.9} & 24.4818 & 22.028 & 23.5806 \\
    \hline
    \textbf{k=1 (learned model only)} & 24.0088 & 21.9016 & 27.2828 \\
    \hline
\end{tabular}
\caption{\label{trajectory length} Trajectory length (per 100 sampling steps).}
\end{table}

\subsection{Chances of Escaping Local Minimum}
One of the motivations behind our work is SA algorithm can only escape local optimum by randomness at the initial high temperature stage. We speculate smoothing the heuristic objective can improve search by eliminating the gaps between nearby local optima in the original objective function. Moreover, the max value function is designed to encourage sacrificing short term gain for long term payoffs. This is done by assigning a high score for states that have a low score, but would later on lead to high scoring states. Hence, we investigate how our models change the frequency of escaping local optima that are in the original objective function.

We first perform SA search with objective function modified by value function, max value function, and Seq2seq model, respectively. Then, the collected search trajectories are re-evaluated by the original objective function to check if any local optima are skipped.
To identify escapes from local optimum, we find all local minima in search trajectories by re-evaluating the original objective score for every three consecutive steps. A step with lower objective score than its two neighbours is considered to be a local minimum. We are specifically interested in whether the search algorithm can learn to sacrifice by stepping into lower score state that later escape to higher score state than even before the local minimum. Hence, only the escapes from local minima that lead to higher score state would be counted.

\begin{table}
\centering
\footnotesize
\begin{tabular}{ |p{3cm}||p{3cm}||p{3cm}||p{3cm}|  }
    \hline
    \textbf{} & \multicolumn{3}{|c|}{\textbf{Number of escapes from local optimum (per 100 steps)}} \\
    \hline
    \textbf{Learned model weight} & \textbf{Seq2seq} & \textbf{Value function} & \textbf{Max value function} \\
    \hline
    \textbf{k=0 (original score)} & 3.5207 & 3.5207 & 3.5207 \\
    \hline
    \textbf{k=0.1} & 3.5132 &  3.3325 & 3.2544 \\
    \hline
    \textbf{k=0.2} & 3.5498 & 3.1663 & 3.1523 \\
    \hline
    \textbf{k=0.3} & 3.4494 & 3.1063 & 3.0562 \\
    \hline
    \textbf{k=0.4} & 3.4241 & 3.1653 & 2.8991 \\
    \hline
    \textbf{k=0.5} & 3.3561 & 3.1203 & 2.7539 \\
    \hline
    \textbf{k=0.6} & 3.2329 & 2.8180 & 2.7269 \\
    \hline
    \textbf{k=0.7} & 3.0843 & 2.7740 & 2.7329 \\
    \hline
    \textbf{k=0.8} & 2.6301 & 2.5537 & 2.2895 \\
    \hline
    \textbf{k=0.9} & 1.6184 & 2.1003 & 1.5397 \\
    \hline
    \textbf{k=1 (learned model only)} & 0.6829 & 1.9566 & 1.3754 \\
    \hline
\end{tabular}
\caption{\label{escape_local} Number of escapes from local optimum (per 100 steps).}
\end{table}

Table \ref{escape_local} presents the number of escapes from local optima per 100 sampling steps for all three proposed models with varying weights. It can be seen that all three models consistently decreases number of escapes from local minimum on average when the weight $k$ increases, which is contrary to our intuition. Perhaps a reasonable explanation would be the learned models learn to identify local minima in the objective function, thus guiding the search algorithm away from even stepping into local minima.

%% file: tex/conclusion.tex
\chapter{Conclusion}

Text generation has been an increasingly trending research area in the field of natural language processing. Natural languages are diverse, complicated, and oftentimes syntactically ambiguous, posing challenge for machines to model natural languages over the years. The emergence of deep neural networks have enabled computer systems to understand, process, and generate complicated natural language. Typically, such deep neural network-based text generation models are trained from parallel corpora by maximizing the likelihood of generating the correct output given the input. Nonetheless, there are numerous scenarios where parallel corpora is not available, calling for unsupervised text generation approach that can generate natural language without the need of parallel supervision. One trending approach to unsupervised text generation is by stochastic search towards a manually designed objective function, which evaluates language fluency, semantic meaning, and other task specific attributes. Such objective function is to be maximized by a local search algorithm that navigates the solution space by starting from the input sentence, then performing word-level or phrase-level edit operations including insertion, replacement, and deletion. Search-based approaches have demonstrated their capability in a variety of tasks, including paraphrase generation, text simplification and keyword-to-text generation.

One of the major drawbacks of search-based text generation models is their performance largely depends on the design of the objective function. Oftentimes, the objective function is heuristically designed to specify desired attributes on a high abstraction level. Such design of objective function is shown to correlate with the true measure of success (BLEU and iBLEU) on a population level, but potentially lack granularity when it comes to each single sentence. Moreover, due to the complex components in the objective function, the optimization landscape is likely to be not smooth, posing significant challenge for the search algorithm to find the optimal solution.

In this dissertation, we address the research question of smoothing and improving the objective function that guides the search. To accomplish this, preliminary search is performed on a given task to collect sample search trajectories. Then we propose three deep neural network-based models to learn and model the search dynamic using the collected search trajectories. Finally, the learned models would be combined with the original objective function to guide a next iteration of search to generate final output.

Experimental results on unsupervised paraphrase generation task with Quora question pairs dataset show all three of our proposed models are indeed capable of improving paraphrase generation performance by adjusting only the objective function. More in-depth analyses show that our three models in fact lead to different search behavior, while all are able to improve the performance in terms BLEU and iBLEU.

Based on what we observed and learned from our study, we identify the follow directions to explore for future work:

\textbf{Iterative Search and Learn}: one direct increment from this dissertation is to have a iterative bootstrapping update between searching and learning. Specifically, one can generate and collect search trajectories using one of our proposed models, which can be used to train a second iteration of value function, max value function or Seq2seq model. The newly learned model can be combined into the objective function again and the process goes on. Previous work \cite{NEURIPS2020_7a677bb4} have shown that a similar bootstrapping alternation between searching and learning using a Seq2seq model to generate new starting points for search can indeed improve search performance over each iteration. It would be interesting to find out if an iterative searching and learning of the search dynamic can leads to improvement in search.
    
\textbf{Adversarial Training}: this idea would only be realistic if a learned model itself suffices for guiding the search, which our models currently don't. However, a sketch of adversarial learning would make sense if such fully learnable objective function exists: the generator in the adversarial framework is simply the search algorithm or the candidate generator, whose goal would be to generate good quality text (e.g. paraphrases); the discriminator plays the role of objective function, whose goal is to differentiate if a search state is acceptable (e.g. if the new state is a good paraphrase). However, the specific training scheme needs to be deliberated to avoid both the generator and discriminator drift too far away from the original objective.
    
\textbf{Embedding Search}: one major limitation of many search-based frameworks is the primitive word-level edit operations: editing text in such a way would require multiple steps to realize large structural modification. However, this could be potentially hard to achieve since most local search algorithms do not explicit retain a history of edits. Hence, if search is performed in the embedding space, one would speculate the underlying sentence structure can change more consistently. The potential challenge of embedding space search is that a tiny shift in numerical embedding space may lead to unchanged discrete text after mapping back to the word space. However, it is still tempting to search by editing a more global representation of text.

%% file: refs.bib
@inproceedings{liu-etal-2020-unsupervised,
    title = "Unsupervised Paraphrasing by Simulated Annealing",
    author = "Liu, Xianggen  and
      Mou, Lili  and
      Meng, Fandong  and
      Zhou, Hao  and
      Zhou, Jie  and
      Song, Sen",
    booktitle = "Proceedings of the 58th Annual Meeting of the Association for Computational Linguistics",
    year = "2020",
    url = "https://www.aclweb.org/anthology/2020.acl-main.28",
    pages = "302--312",
}

@inproceedings{Miao_Zhou_Mou_Yan_Li_2019,
    title={{CGMH}: Constrained Sentence Generation by Metropolis-Hastings Sampling}, 
    url={https://ojs.aaai.org/index.php/AAAI/article/view/4659}, 
    booktitle={Proceedings of the Thirty-Third AAAI Conference on Artificial Intelligence}, 
    author={Miao, Ning and Zhou, Hao and Mou, Lili and Yan, Rui and Li, Lei}, 
    year={2019}, 
    pages={6834-6842} }

@inproceedings{devlin-etal-2019-bert,
    title = "{BERT}: Pre-training of Deep Bidirectional Transformers for Language Understanding",
    author = "Devlin, Jacob  and
      Chang, Ming-Wei  and
      Lee, Kenton  and
      Toutanova, Kristina",
    booktitle = "Proceedings of the 2019 Conference of the North {A}merican Chapter of the Association for Computational Linguistics: Human Language Technologies",
    year = "2019",
    url = "https://www.aclweb.org/anthology/N19-1423",
    pages = "4171--4186",
}

@inproceedings{dong-etal-2019-editnts,
    title = "{E}dit{NTS}: An Neural Programmer-Interpreter Model for Sentence Simplification through Explicit Editing",
    author = "Dong, Yue  and
      Li, Zichao  and
      Rezagholizadeh, Mehdi  and
      Cheung, Jackie Chi Kit",
    booktitle = "Proceedings of the 57th Annual Meeting of the Association for Computational Linguistics",
    year = "2019",
    url = "https://www.aclweb.org/anthology/P19-1331",
    pages = "3393--3402",
}

@inproceedings{NEURIPS2019_675f9820,
 author = {Gu, Jiatao and Wang, Changhan and Zhao, Junbo},
 booktitle = {Advances in Neural Information Processing Systems},
 title = {Levenshtein Transformer},
 url = {shorturl.at/wQ123},
 year = {2019}
}

@inproceedings{HC,
    title = "Discrete Optimization for Unsupervised Sentence Summarization with Word-Level Extraction",
    author = "Schumann, Raphael and Mou, Lili and Lu, Yao and Vechtomova, Olga and Markert, Katja",
    booktitle = "Proceedings of the 58th Annual Meeting of the Association for Computational Linguistics",
    year = "2020",
    pages = "5032--5042",
    url= "https://www.aclweb.org/anthology/2020.acl-main.452/"
}

@inproceedings{transformer,
    author = {Vaswani, Ashish and Shazeer, Noam and Parmar, Niki and Uszkoreit, Jakob and Jones, Llion and Gomez, Aidan N and Kaiser, \L ukasz and Polosukhin, Illia},
    booktitle = {Advances in Neural Information Processing Systems},
    pages = {},
    title = {Attention is All you Need},
    url = {shorturl.at/dBGX0},
    year = {2017}
}

@inproceedings{kumar-etal-2020-iterative,
    title = "Iterative Edit-Based Unsupervised Sentence Simplification",
    author = "Kumar, Dhruv  and
      Mou, Lili  and
      Golab, Lukasz  and
      Vechtomova, Olga",
    booktitle = "Proceedings of the 58th Annual Meeting of the Association for Computational Linguistics",
    year = "2020",
    url = "https://www.aclweb.org/anthology/2020.acl-main.707",
    pages = "7918--7928",
}

@inproceedings{NEURIPS2020_7a677bb4,
     author = {Li, Jingjing and Li, Zichao and Mou, Lili and Jiang, Xin and Lyu, Michael and King, Irwin},
     booktitle = {Advances in Neural Information Processing Systems},
     pages = {10820--10831},
     title = {Unsupervised Text Generation by Learning from Search},
     url = {shorturl.at/djmpP},
     year = {2020}
}

@inproceedings{bowman-etal-2016-generating,
    title = "Generating Sentences from a Continuous Space",
    author = "Bowman, Samuel R.  and
      Vilnis, Luke  and
      Vinyals, Oriol  and
      Dai, Andrew  and
      Jozefowicz, Rafal  and
      Bengio, Samy",
    booktitle = "Proceedings of the 20th {SIGNLL} Conference on Computational Natural Language Learning",
    year = "2016",
    url = "https://www.aclweb.org/anthology/K16-1002",
}

@inproceedings{papineni-etal-2002-bleu,
    title = "{BLEU}: a Method for Automatic Evaluation of Machine Translation",
    author = "Papineni, Kishore  and
      Roukos, Salim  and
      Ward, Todd  and
      Zhu, Wei-Jing",
    booktitle = "Proceedings of the 40th Annual Meeting of the Association for Computational Linguistics",
    year = "2002",
    url = "https://www.aclweb.org/anthology/P02-1040",
    pages = "311--318",
}

@book{lyons1991natural,
  title={Natural Language and Universal Grammar: Volume 1: Essays in Linguistic Theory},
  author={Lyons, John},
  year={1991},
  publisher={Cambridge University Press}
}

@inproceedings{malmi-etal-2019-encode,
    title = "Encode, Tag, Realize: High-Precision Text Editing",
    author = "Malmi, Eric  and
      Krause, Sebastian  and
      Rothe, Sascha  and
      Mirylenka, Daniil  and
      Severyn, Aliaksei",
    booktitle = "Proceedings of the 2019 Conference on Empirical Methods in Natural Language Processing and the 9th International Joint Conference on Natural Language Processing",
    year = "2019",
    url = "https://www.aclweb.org/anthology/D19-1510",
    pages = "5054--5065",
}

@inproceedings{li-etal-2018-paraphrase,
    title = "Paraphrase Generation with Deep Reinforcement Learning",
    author = "Li, Zichao  and
      Jiang, Xin  and
      Shang, Lifeng  and
      Li, Hang",
    booktitle = "Proceedings of the 2018 Conference on Empirical Methods in Natural Language Processing",
    year = "2018",
    url = "https://www.aclweb.org/anthology/D18-1421",
    pages = "3865--3878",
}

@inproceedings{NIPS2014_a14ac55a,
    author = {Sutskever, Ilya and Vinyals, Oriol and Le, Quoc V},
    booktitle = {Advances in Neural Information Processing Systems},
    pages = {},
    title = {Sequence to Sequence Learning with Neural Networks},
    url = {https://proceedings.neurips.cc/paper/2014/file/a14ac55a4f27472c5d894ec1c3c743d2-Paper.pdf},
    year = {2014}
}

@article{rose2010automatic,
  title={Automatic keyword extraction from individual documents},
  author={Rose, Stuart and Engel, Dave and Cramer, Nick and Cowley, Wendy},
  journal={Text Mining: Applications and Theory},
  pages={1--20},
  year={2010},
  publisher={Citeseer},
  url="https://onlinelibrary.wiley.com/doi/abs/10.1002/9780470689646.ch1"
}

@article{rl_value_function,
author = {Boyan, Justin and Moore, Andrew W.},
title = {Learning Evaluation Functions to Improve Optimization by Local Search},
year = {2001},
issue_date = {9/1/2001},
publisher = {JMLR.org},
issn = {1532-4435},
url = {https://dl.acm.org/doi/10.1162/15324430152733124},
journal = {Journal of Machine Learning Research},
pages = {77–112},
numpages = {36}
}

@inproceedings{gu-etal-2016-incorporating,
    title = "Incorporating Copying Mechanism in Sequence-to-Sequence Learning",
    author = "Gu, Jiatao  and
      Lu, Zhengdong  and
      Li, Hang  and
      Li, Victor O.K.",
    booktitle = "Proceedings of the 54th Annual Meeting of the Association for Computational Linguistics",
    year = "2016",
    url = "https://aclanthology.org/P16-1154",
}

@inproceedings{xia-etal-2020-bert,
    title = "Which *{BERT}? {A} Survey Organizing Contextualized Encoders",
    author = "Xia, Patrick  and
      Wu, Shijie  and
      Van Durme, Benjamin",
    booktitle = "Proceedings of the 2020 Conference on Empirical Methods in Natural Language Processing",
    year = "2020",
    url = "https://aclanthology.org/2020.emnlp-main.608",
}

@inproceedings{sun-zhou-2012-joint,
    title = "Joint Learning of a Dual {SMT} System for Paraphrase Generation",
    author = "Sun, Hong  and
      Zhou, Ming",
    booktitle = "Proceedings of the 50th Annual Meeting of the Association for Computational Linguistics (Volume 2: Short Papers)",
    year = "2012",
    url = "https://aclanthology.org/P12-2008",
}

@inproceedings{alva-manchego-etal-2017-learning,
    title = "Learning How to Simplify From Explicit Labeling of Complex-Simplified Text Pairs",
    author = "Alva-Manchego, Fernando  and
      Bingel, Joachim  and
      Paetzold, Gustavo  and
      Scarton, Carolina  and
      Specia, Lucia",
    booktitle = "Proceedings of the Eighth International Joint Conference on Natural Language Processing (Volume 1: Long Papers)",
    year = "2017",
    url = "https://aclanthology.org/I17-1030",
}

@article{kingma2013auto,
  title={Auto-encoding variational bayes},
  author={Kingma, Diederik P and Welling, Max},
  journal={arXiv preprint arXiv:1312.6114},
  year={2013}
}

@article{goldberg1994using,
  title={Using natural-language processing to produce weather forecasts},
  author={Goldberg, Eli and Driedger, Norbert and Kittredge, Richard I},
  journal={IEEE Expert},
  volume={9},
  number={2},
  pages={45--53},
  year={1994},
  publisher={IEEE}
}

@book{goodfellow2016deep,
  title={Deep learning},
  author={Goodfellow, Ian and Bengio, Yoshua and Courville, Aaron},
  year={2016},
  publisher={MIT press}
}

@article{word2vec,
  title={Efficient estimation of word representations in vector space},
  author={Mikolov, Tomas and Chen, Kai and Corrado, Greg and Dean, Jeffrey},
  journal={arXiv preprint arXiv:1301.3781},
  year={2013}
}

@inproceedings{pennington2014glove,
  title={Glove: Global vectors for word representation},
  author={Pennington, Jeffrey and Socher, Richard and Manning, Christopher D},
  booktitle={Proceedings of the 2014 conference on empirical methods in natural language processing (EMNLP)},
  pages={1532--1543},
  year={2014}
}

@article{bahdanau2014neural,
  title={Neural machine translation by jointly learning to align and translate},
  author={Bahdanau, Dzmitry and Cho, Kyunghyun and Bengio, Yoshua},
  journal={arXiv preprint arXiv:1409.0473},
  year={2014}
}

@inproceedings{zhu2015aligning,
  title={Aligning books and movies: Towards story-like visual explanations by watching movies and reading books},
  author={Zhu, Yukun and Kiros, Ryan and Zemel, Rich and Salakhutdinov, Ruslan and Urtasun, Raquel and Torralba, Antonio and Fidler, Sanja},
  booktitle={Proceedings of the IEEE international conference on computer vision},
  pages={19--27},
  year={2015}
}

@article{rogers2020primer,
  title={A primer in bertology: What we know about how bert works},
  author={Rogers, Anna and Kovaleva, Olga and Rumshisky, Anna},
  journal={Transactions of the Association for Computational Linguistics},
  year={2020},
}

@article{gpt,
  title={Improving language understanding by generative pre-training},
  author={Radford, Alec and Narasimhan, Karthik and Salimans, Tim and Sutskever, Ilya},
  year={2018}
}

@article{gpt2,
  title={Language models are unsupervised multitask learners},
  author={Radford, Alec and Wu, Jeffrey and Child, Rewon and Luan, David and Amodei, Dario and Sutskever, Ilya and others},
  journal={OpenAI blog},
  volume={1},
  number={8},
  pages={9},
  year={2019}
}

@inproceedings{zhao-etal-2018-integrating,
    title = "Integrating Transformer and Paraphrase Rules for Sentence Simplification",
    author = "Zhao, Sanqiang  and
      Meng, Rui  and
      He, Daqing  and
      Saptono, Andi  and
      Parmanto, Bambang",
    booktitle = "Proceedings of the 2018 Conference on Empirical Methods in Natural Language Processing",
    year = "2018",
    publisher = "Association for Computational Linguistics",
    url = "https://aclanthology.org/D18-1355",
}

@article{zellers2018swag,
  title={Swag: A large-scale adversarial dataset for grounded commonsense inference},
  author={Zellers, Rowan and Bisk, Yonatan and Schwartz, Roy and Choi, Yejin},
  journal={arXiv preprint arXiv:1808.05326},
  year={2018}
}

@InProceedings{wiki_dataset,
  author = 	"Zhang, Xingxing
		and Lapata, Mirella",
  title = 	"Sentence Simplification with Deep Reinforcement Learning",
  booktitle = 	"Proceedings of the 2017 Conference on Empirical Methods in Natural Language Processing",
  year = 	"2017",
  publisher = 	"Association for Computational Linguistics",
  url = 	"http://aclweb.org/anthology/D17-1063"
}

@inproceedings{ballard1987modular,
  title={Modular learning in neural networks.},
  author={Ballard, Dana H},
  booktitle={AAAI},
  volume={647},
  pages={279--284},
  year={1987}
}

@inproceedings{kiros2015skip,
  title={Skip-thought vectors},
  author={Kiros, Ryan and Zhu, Yukun and Salakhutdinov, Russ R and Zemel, Richard and Urtasun, Raquel and Torralba, Antonio and Fidler, Sanja},
  booktitle={Advances in neural information processing systems},
  pages={3294--3302},
  year={2015}
}

@inproceedings{pagliardini-etal-2018-unsupervised,
    title = "Unsupervised Learning of Sentence Embeddings Using Compositional n-Gram Features",
    author = "Pagliardini, Matteo  and
      Gupta, Prakhar  and
      Jaggi, Martin",
    booktitle = "Proceedings of the 2018 Conference of the North {A}merican Chapter of the Association for Computational Linguistics: Human Language Technologies, Volume 1 (Long Papers)",
    year = "2018",
    url = "https://aclanthology.org/N18-1049",
}

@inproceedings{kann-etal-2018-sentence,
    title = "Sentence-Level Fluency Evaluation: References Help, But Can Be Spared!",
    author = "Kann, Katharina  and
      Rothe, Sascha  and
      Filippova, Katja",
    booktitle = "Proceedings of the 22nd Conference on Computational Natural Language Learning",
    year = "2018",
    url = "https://aclanthology.org/K18-1031",
}

@techreport{kincaid1975derivation,
  title={Derivation of new readability formulas (automated readability index, fog count and flesch reading ease formula) for navy enlisted personnel},
  author={Kincaid, J Peter and Fishburne Jr, Robert P and Rogers, Richard L and Chissom, Brad S},
  year={1975},
  institution={Naval Technical Training Command Millington TN Research Branch}
}

@inproceedings{luong-etal-2015-effective,
    title = "Effective Approaches to Attention-based Neural Machine Translation",
    author = "Luong, Thang  and
      Pham, Hieu  and
      Manning, Christopher D.",
    booktitle = "Proceedings of the 2015 Conference on Empirical Methods in Natural Language Processing",
    year = "2015",
    url = "https://aclanthology.org/D15-1166",
}

@article{wu2016google,
  title={Google's neural machine translation system: Bridging the gap between human and machine translation},
  author={Wu, Yonghui and Schuster, Mike and Chen, Zhifeng and Le, Quoc V and Norouzi, Mohammad and Macherey, Wolfgang and Krikun, Maxim and Cao, Yuan and Gao, Qin and Macherey, Klaus and others},
  journal={arXiv preprint arXiv:1609.08144},
  year={2016}
}

@inproceedings{wiseman-etal-2017-challenges,
    title = "Challenges in Data-to-Document Generation",
    author = "Wiseman, Sam  and
      Shieber, Stuart  and
      Rush, Alexander",
    booktitle = "Proceedings of the 2017 Conference on Empirical Methods in Natural Language Processing",
    year = "2017",
    url = "https://aclanthology.org/D17-1239",
}

@inproceedings{puduppully2019data,
  title={Data-to-text generation with content selection and planning},
  author={Puduppully, Ratish and Dong, Li and Lapata, Mirella},
  booktitle={Proceedings of the AAAI conference on artificial intelligence},
  volume={33},
  number={01},
  pages={6908--6915},
  year={2019}
}

@inproceedings{la-quatra-cagliero-2020-end,
    title = "End-to-end Training For Financial Report Summarization",
    author = "La Quatra, Moreno  and
      Cagliero, Luca",
    booktitle = "Proceedings of the 1st Joint Workshop on Financial Narrative Processing and MultiLing Financial Summarisation",
    year = "2020",
    url = "https://aclanthology.org/2020.fnp-1.20",
}

@inproceedings{el-haj-etal-2020-financial,
    title = "The Financial Narrative Summarisation Shared Task ({FNS} 2020)",
    author = "El-Haj, Mahmoud  and
      AbuRa{'}ed, Ahmed  and
      Litvak, Marina  and
      Pittaras, Nikiforos  and
      Giannakopoulos, George",
    booktitle = "Proceedings of the 1st Joint Workshop on Financial Narrative Processing and MultiLing Financial Summarisation",
    year = "2020",
}

@article{hornik1989multilayer,
  title={Multilayer feedforward networks are universal approximators},
  author={Hornik, Kurt and Stinchcombe, Maxwell and White, Halbert},
  journal={Neural networks},
  volume={2},
  number={5},
  pages={359--366},
  year={1989},
  publisher={Elsevier}
}

@misc{havens2019fitbert,
    title  = {Use BERT to Fill in the Blanks},
    author = {Sam Havens and Aneta Stal},
    url    = {https://github.com/Qordobacode/fitbert},
    year   = {2019}
}

@article{hochreiter1997long,
  title={Long short-term memory},
  author={Hochreiter, Sepp and Schmidhuber, J{\"u}rgen},
  journal={Neural computation},
  volume={9},
  number={8},
  pages={1735--1780},
  year={1997},
  publisher={MIT Press}
}

@article{hochreiter1998vanishing,
  title={The vanishing gradient problem during learning recurrent neural nets and problem solutions},
  author={Hochreiter, Sepp},
  journal={International Journal of Uncertainty, Fuzziness and Knowledge-Based Systems},
  volume={6},
  number={02},
  pages={107--116},
  year={1998},
  publisher={World Scientific}
}

@inproceedings{jozefowicz2015empirical,
  title={An empirical exploration of recurrent network architectures},
  author={Jozefowicz, Rafal and Zaremba, Wojciech and Sutskever, Ilya},
  booktitle={International conference on machine learning},
  pages={2342--2350},
  year={2015},
  organization={PMLR}
}
